\documentclass[12pt]{iopart}

\usepackage{geometry}
\usepackage[T1]{fontenc}
\usepackage{lmodern}
\usepackage{iopams} 
\usepackage[symbol]{footmisc} 
\usepackage{import}
\usepackage{iopams} 
\usepackage[symbol]{footmisc}
\usepackage[utf8]{inputenc} 
\usepackage[T1]{fontenc}    
\usepackage{hyperref}       
\usepackage{booktabs}       
\usepackage{nicefrac}       
\usepackage{microtype}      
\usepackage{enumitem}
\usepackage{wrapfig}
\usepackage{cancel}
\usepackage{pgfplots}
\usepackage{pdfpages}
\usepackage{listliketab}
\usepackage{graphicx}
\usepackage{amsmath,amssymb,amsthm,amscd,amsfonts,amsbsy}
\usepackage{mathtools}
\usepackage{url}
\usepackage{verbatim}
\usepackage{slashed}
\usepackage{bbm}
\usepackage{algorithm}
\usepackage{algorithmicx}
\usepackage{algpseudocode}
\usepackage{array}
\usepackage{csquotes}
\usepackage{braket}
\usepackage{float}
\usepackage[bf]{caption} 
\usepackage{adjustbox}
\usepackage{soul}
\usepackage{subcaption}

\usepackage[
backend=bibtex,
sorting=none,
]{biblatex}
\usepackage{tabularray}

\algdef{SxNE}[DOBLOCK]{Do}{EndDoBlock}[1]{\algorithmicdo\ #1}{}%

\newcommand{\bitem}{\begin{list}{$\bullet$}%
{\setlength{\itemsep}{0pt}\setlength{\topsep}{0pt}%
\setlength{\rightmargin}{0pt}}} \newcommand{\eitem}{\end{list}}

\DeclareMathOperator*{\argmin}{arg\,min}

\theoremstyle{theorem}
\newtheorem{thm}{Theorem}[section]
\newtheorem{lem}[thm]{Lemma}
\newtheorem{prop}[thm]{Proposition}
\newtheorem{cor}[thm]{Corollary}

\theoremstyle{definition}
\newtheorem{dfn}[thm]{Definition}

\theoremstyle{remark}
\newtheorem*{rem}{Remark}

\newcommand{\bdfn}[1][]{\ifthenelse{\equal{#1}{}}{\begin{dfn}}{\begin{dfn}[#1]}}
\newcommand{\edfn}{\end{dfn}}
\newcommand{\bthm}[1][]{\ifthenelse{\equal{#1}{}}{\begin{thm}}{\begin{thm}[#1]}}
\newcommand{\ethm}{\end{thm}}
\newcommand{\bcor}[1][]{\ifthenelse{\equal{#1}{}}{\begin{cor}}{\begin{cor}[#1]}}
\newcommand{\ecor}{\end{cor}}
\newcommand{\blem}[1][]{\ifthenelse{\equal{#1}{}}{\begin{lem}}{\begin{lem}[#1]}}
\newcommand{\elem}{\end{lem}}
\newcommand{\bprop}[1][]{\ifthenelse{\equal{#1}{}}{\begin{prop}}{\begin{prop}[#1]}}
\newcommand{\eprop}{\end{prop}}
\newcommand{\brem}{\begin{rem}}
\newcommand{\erem}{\end{rem}}
\newcommand\bpf[1][]{
\ifthenelse{\equal{#1}{}}{
\begin{proof}}
{\begin{proof}(#1)}
}
\newcommand\epf{\end{proof}}

\newcommand{\eat}[1]{}

\newcommand{\zstroke}{%
  \text{\ooalign{\hidewidth -\kern-.3em-\hidewidth\cr$z$\cr}}%
}

\newcommand\norm[1]{\left\lVert #1 \right\rVert}

\makeatletter
\newcommand{\iid}[0]{\mathbin{\overset{iid}{\kern\z@\sim}}}

\makeatletter
\DeclareRobustCommand*{\bfseries}{%
  \not@math@alphabet\bfseries\mathbf
  \fontseries\bfdefault\selectfont
  \boldmath
}
\makeatother

\begin{document}

\title[]{Quasiprobabilistic Density Ratio Estimation with a Reverse Engineered Classification Loss Function}
\author{Matthew Drnevich*$^a$, Stephen Jiggins$^b$, Kyle Cranmer$^c$}
\address{$^a$ Physics Department, New York University, \\ \; EMail: mdd424@nyu.edu}
\address{$^b$ Deutsches Elektronen-Synchrotron DESY, Germany, \\ \; EMail: stephen.jiggins@desy.de}
\address{$^c$ Physics Department, University of Wisconsin--Madison, \\ \; Data Science Institute, University of Wisconsin--Madison, \\ \; EMail: kyle.cranmer@wisc.edu}
\def\thefootnote{*}\footnotetext{Denotes primary contributor}\def\thefootnote{\arabic{footnote}}

\vspace{10pt}
\begin{indented}
\item[]\today
\end{indented}

\begin{abstract}

We consider a generalization of the classifier-based density-ratio estimation task to a quasiprobabilistic setting where probability densities can be negative. The problem with most loss functions used for this task is that they implicitly define a relationship between the optimal classifier and the target quasiprobabilistic density ratio which is discontinuous or not surjective. We address these problems by introducing a convex loss function that is well-suited for both probabilistic and quasiprobabilistic density ratio estimation. To quantify performance, an extended version of the Sliced-Wasserstein distance is introduced which is compatible with quasiprobability distributions. We demonstrate our approach on a real-world example from particle physics, of di-Higgs production in association with jets via gluon-gluon fusion, and achieve state-of-the-art results.

\end{abstract}

\section{Introduction}

In many domains of science, density ratios---also known as likelihood ratios---play an important role. They are central to hypothesis tests~\cite{NeymanPearsonLemma}, importance sampling~\cite{Lemi09}, information theory~\cite{song2020understandinglimitationsvariationalmutual}, and domain adaptation~\cite{NIPS2011_d1f255a3}. In the context of machine learning, the probability densities associated to data distributions are typically unknown and in scientific settings probability densities are often defined implicitly through a complex simulator~\cite{MC_Implicit_Models}.
For example, in High Energy Physics (HEP), Monte Carlo based simulators are heavily utilised to generate synthetic data associated to various theoretical scenarios~\cite{MonteCarloChallenges-2021}.

Consequently, neural density ratio estimation (NDRE) has seen wide spread adoption in an array of tasks from simulation-based inference~\cite{cranmer2020frontier,HZZ_ATLAS_2025} to inverse unfolding problems~\cite{ijcai2023p162,omnifold}. Perhaps the most common approach to NDRE is to train a (probabilistic) binary classifier, and then transform its output so that it serves as an estimate for the density ratio, through what is colloquially referred to as the ``likelihood ratio trick''~\cite{cranmer2016approximating}. The aforementioned transformation is based on the relationship between the Bayes optimal classifier and the targeted density ratio. In this work, we will pursue a generalization of this strategy, in which the function that optimizes a modified loss function is related to the target density ratio through some other transformation. 

Our motivation for pursuing this alternative strategy is the need to model quasiprobabilistic density ratios, which can be negative. The quasiprobabilistic setting, in which the first Kolmogorov axiom is violated~\cite{ANKolmogorov1933}, arises naturally when modeling quantum mechanical systems, either due to inherently non-classical phenomena such as quantum interference, or as a consequence of approximation techniques employed to describe particle collisions~\cite{Frederix_2020,danziger2021reducing,Nason:2004rx,Frixione:2007vw}. The presence of negative densities poses two immediate problems for the vanilla (probabilistic) likelihood ratio trick. First, the output of standard binary classifiers is constrained to be positive, typically as a result of activation functions such as the sigmoid. Second, the family of loss functions used to train binary classifiers~\cite{Rizvi_2024}, such as binary cross-entropy, are incompatible with negative densities.

Extending neural density ratio estimation to a quasiprobabilistic setting has been demonstrated via the work of Ref.~\cite{drnevich2024neuralquasiprobabilisticlikelihoodratio}, in which signed probabilistic measures~\cite{Interpretations_of_Probability}, and a quasiprobabilistic-safe loss function, were introduced. In this work, we focus on introducing an analogue of the ``likelihood ratio trick'' for quasiprobabilsitic classifier-based density ratio estimation through the introduction of a new convex loss function. This approach addresses some of the shortcomings of the \texttt{PARE} loss function introduced in Ref.~\cite{drnevich2024neuralquasiprobabilisticlikelihoodratio} and allows one to train a more unrestricted class of models than was previously possible.

The paper is organized as follows. In Section \ref{sec:Losses}, the challenge of applying classifier-based neural density ratio estimation, when faced with learning quasiprobabilistic density ratios, is outlined. A solution to this problem is presented in the form of a new quasiprobabilistic ratio trick, culminating in a novel loss function referred to as the \textbf{REV}erse \textbf{E}ngineered \textbf{R}atio \textbf{T}rick (REVERT) loss. In Section \ref{sec:QP-SMEFT}, we demonstrate this new approach on a real world example found in particle physics, by comparing the application of the new loss function to a simple Multilayer Perceptron (dense feed-forward neural network) classifier, in addition to the optimization problem of training the Ratio of Signed Mixtures Models (RoSMM) found in Ref.~\cite{drnevich2024neuralquasiprobabilisticlikelihoodratio}. Finally, we conclude with closing remarks in Section~\ref{sec:Conclusions}.
\label{sec:Introduction}

\section{Density Ratio Estimation}
\label{sec:Losses}
The problem addressed by this paper is that of learning the density ratio between two quasiprobabilistic distributions using neural classifier-based methods given by:
\begin{align}
\label{eq:QLR}
    r^*(\mathbf{x}) \equiv \frac{q(\mathbf{x}|Y=1)}{q(\mathbf{x}|Y=0)},
\end{align}
where the random variable $Y$ is a class label designating from which distribution a dataset is sampled, taking values in $\{0,1\}$. The distribution $q(\cdot)$ is used to differentiate the quasiprobabilistic density function from a probabilistic density function $p(\cdot)$. As such, it is allowed that $q(\mathbf{x}|y) < 0$ for some $\mathbf{x}$. The relationship between the optimal classifier $s^*$ for a given loss function $\mathcal{L}(s)$ and the density ratio $r^*$ defines the aforementioned ``ratio trick''. For the class of loss functions considered herein, the resulting optimised classifier:
\begin{equation}
    s^{*} = \argmin_{s} \, \mathcal{L}(s),
\end{equation}
is directly related to the density ratio $r^*$ through some transformation $T$, such that $r^* = T(s^*)$ \footnote{Note that $T$ here is the inverse of the convention used in Ref. \cite{drnevich2024neuralquasiprobabilisticlikelihoodratio}, but this is the more natural form to use in this setup.}. 

\begin{figure}
    \centering
    \includegraphics[width=0.75\linewidth]{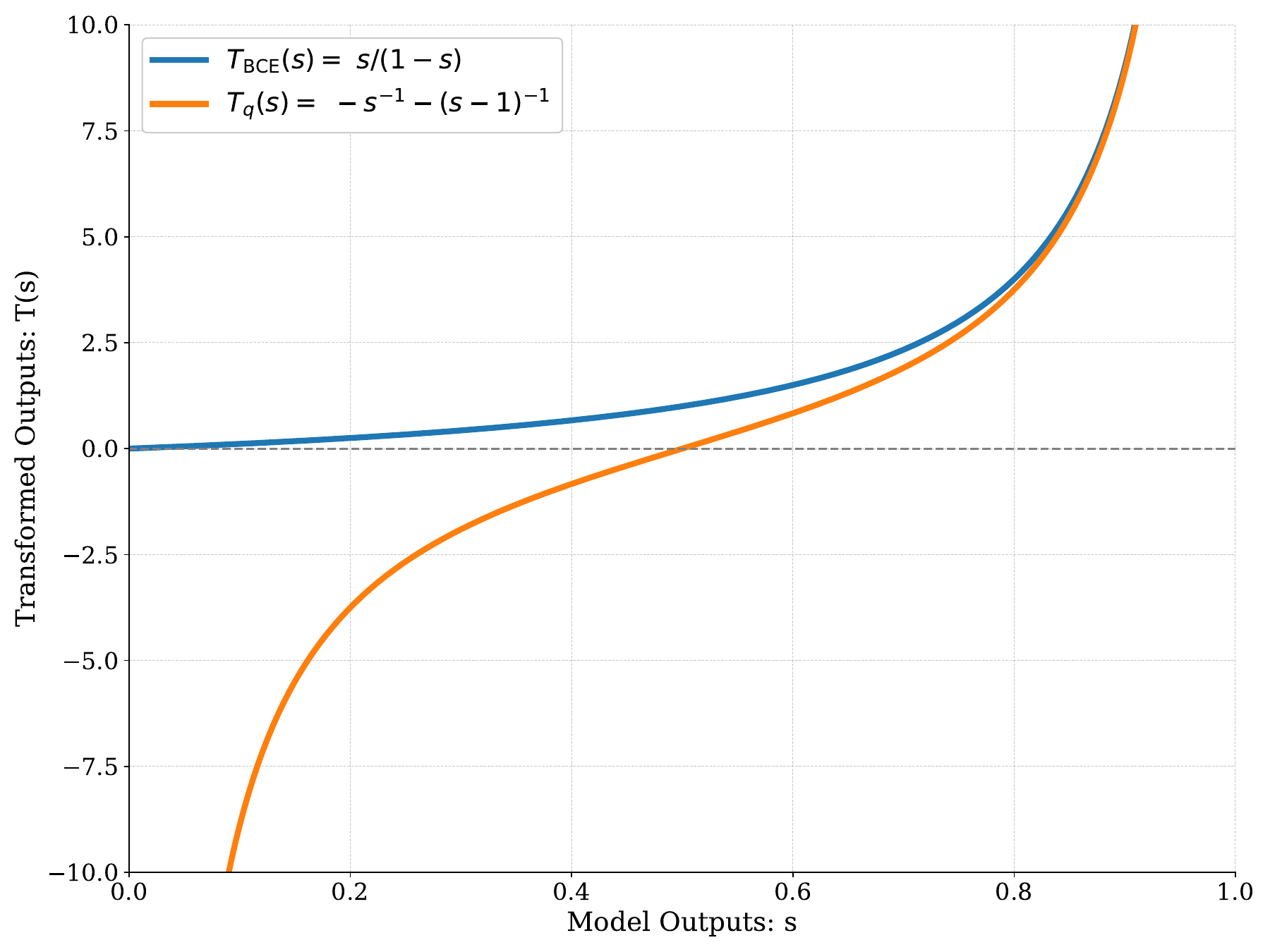}
    \caption{Examples of transformations between the output space of a classifier $s$ which outputs values in the interval $(0,1)$ and the transformed output of the classifier using ``ratio trick'' transformations. $T_{\rm{BCE}}$ is the transformation associated with the binary cross-entropy loss function and $T_q$ is an example of a transformation suitable for quasiprobabilistic settings. $T_{q}$ can transform the classifier outputs to any real number, while $T_{\rm{BCE}}$ transforms all classifier outputs to positive numbers only. 
    }
    \label{fig:Domain-Codomain}
\end{figure}

A key characteristic of these transformations is that they should be homeomorphisms (continuous bijective functions) between the codomain of the optimizable function, $s$ (the space of the model outputs), and the codomain of the density ratio function, $r$ (the set of possible density ratio values). In the probabilistic setting, the density ratio is always nonnegative, while a quasiprobabilistic density ratio can take on any real value. Accordingly, standard probabilistic density ratio estimation loss functions are constructed such that their associated transformations are homeomorphisms between the space of classifier outputs and the positive real line $\mathbb{R}_{>0}$. As a result, these transformations are unable to map any of the classifier outputs to negative density ratio values, even if the data is quasiprobabilistic; as depicted by Figure \ref{fig:Domain-Codomain}.
This motivates the construction of a new loss function for quasiprobabilistic settings where the associated ratio trick transformation is a homeomorphism between the space of model outputs and all real numbers.


\subsection{Deriving a Loss Function}
One way to construct a suitable loss function is by starting with a ratio trick transformation, and then reverse engineering a loss function that gives rise to said transformation. To achieve this, we first consider how the ratio trick~\cite{cranmer2016approximating} process works for binary classification. For a generic loss function $\mathcal{L}$, the risk $(R)$ is given by:
\begin{equation}\label{eq:empirical_risk}
    R[s] = \mathbb{E}_{X,Y}\left[\mathcal{L}(s(X),Y)\right] = \mathbb{E}_{Y}\mathbb{E}_{X|Y}\left[\mathcal{L}(s(X),Y)\right] = \int_{\mathcal{X}} \underbrace{\sum_{y\in\{0,1\}}\mathcal{L}(s(\mathbf{x}),y)q(\mathbf{x}|y)p(y)}_{{\textrm{Lagrangian:} \; L(\mathbf{x},s)}}d\mathbf{x},
\end{equation}
where the integrand in Eq. \ref{eq:empirical_risk} is known as a Lagrangian $L$. We consider a common form of loss functions for binary classification where:
\begin{equation}\label{eq:classification_loss}
    \mathcal{L}(s,y) = yf(s) + (1-y)g(s)
\end{equation}
for two functions $f,g: \mathrm{cod(s)} \to \mathbb{R}$ \footnote{We use the shorthand notation $\mathrm{cod}(s)$ for $\mathrm{codomain}(s)$}. Consequently, the Lagrangian can be written as:
\begin{equation}\label{eq:lagrangian}\begin{split}
    L(\mathbf{x},s) &= \sum_{y\in\{0,1\}}\mathcal{L}(s,y)q(\mathbf{x}|y)p(y) \\
    &= g(s)q(\mathbf{x}|Y=0)p(Y=0) + f(s)q(\mathbf{x}|Y=1)p(Y=1)
\end{split}
\end{equation}
The extrema of the risk can then be found by using the Euler-Lagrange equation, which states that if the function $s^{*}$ is an extremum then:
\begin{equation}\label{eq:EL_solution}
    \left. \frac{\partial L}{\partial s}\right|_{s=s^{*}(\mathbf{x})} = 0 \quad\Longrightarrow\quad \frac{q(\mathbf{x}|Y=1)p(Y=1)}{q(\mathbf{x}|Y=0)p(Y=0)} = -\frac{g'(s^*(\mathbf{x}))}{f'(s^*(\mathbf{x}))},
\end{equation}
where the superscript $'$ denotes the derivative with respect to the classifier function $s$; e.g. $g'(s) = d g/ds$. For balanced datasets with $p(Y=0) = p(Y=1)$, the right-hand side equation in Eq. \ref{eq:EL_solution} combined with Eq. \ref{eq:QLR} becomes:
\begin{equation}
    r^*(\mathbf{x}) = -\frac{g'(s^*(\mathbf{x}))}{f'(s^*(\mathbf{x}))} 
\end{equation}
which allows us to define the transformation $T$ as the ratio $-g'/f'$ such that~\footnote{A full derivation of this ratio trick can be found in Appendix \ref{app:LossDerivation}}:
\begin{equation}\label{eq:RatioTrick}
    T(s^*(\mathbf{x})) = -\frac{g'(s^*(\mathbf{x}))}{f'(s^*(\mathbf{x}))} = r^*(\mathbf{x})
\end{equation}

This work focuses on a particular class of such loss functions which have desirable properties. All binary classification loss functions with $f(s)=s$ are considered:
\begin{equation}\label{eq:loss_with_g}
    \mathcal{L}(s,y) = ys + (1-y)g(s),
\end{equation}
because of the following properties. For this class of loss functions, the convexity of the Lagrangian in Eq. \ref{eq:lagrangian} with respect to $s$ is ensured, up to a change in sign for $g$, provided that $q(\mathbf{x}|Y=0)$ is nonnegative and the transformation given by Eq.~\ref{eq:RatioTrick} is a homeomorphism. Nonnegativity of $q(\mathbf{x}|Y=0)$ is a mild requirement since if $q(\mathbf{x}|Y=0) < 0$ at any point then it must cross zero at some point, rendering the ratio in Eq.~\ref{eq:QLR} undefined at that point. In this situation, density ratio estimation becomes an ill-posed learning problem since the ratio is unbounded and discontinuous. Most importantly, convexity of the integrand in Eq. \ref{eq:empirical_risk} with respect to $s$ ensures that the Euler-Lagrange solution given by Eq. \ref{eq:RatioTrick} is the unique minimizer of the risk. Formal derivations of these properties are provided in Appendix~\ref{app:Convexity}.

Finally, with the construction in Eq. \ref{eq:loss_with_g}, it follows from Eq. \ref{eq:RatioTrick} that $g(s)$ is the anti-derivative of $-T(s)$:
\begin{equation}\label{eq:diff_equation}
    g(s) = \int -T(s)ds.
\end{equation}
Therefore, the loss function can be written in terms of the ratio trick transformation:
\begin{equation}\label{eq:new_loss_formula}
    \mathcal{L}(s,y) = ys - (1-y)\int T(s)ds
\end{equation}
The reverse engineering strategy presented herein is to choose an appropriate transformation $T$ which subsequently defines the loss function $\mathcal{L}$.

\subsection{Reverse Engineered Ratio Trick Loss}

To construct a loss function suitable for quasiprobabilistic density ratio, we start by choosing a ratio trick transformation that is a homeomorphism between the space of model outputs (the codomain of $s$) and the space of all real numbers $\mathbb{R}$. Once chosen, Eq.~\ref{eq:new_loss_formula} allows us to define the associated loss function which will necessarily be convex (up to a change in sign). For numerical optimization, it is also beneficial that the model outputs be bounded, so the possible outputs of the model are restricted to some finite interval $(a,b)$ with $a<b$. 
Therefore, we choose a homeomorphism $T:(a,b) \to \mathbb{R}$ to define the ratio trick transformation. One such transformation is:
\begin{equation}
    T(s;a,b) = \frac{1}{s-a} + \frac{1}{s-b},
    \label{eq:REVERT_Ratio_Trick_General_Form}
\end{equation}
which corresponds to the loss function
\begin{equation}
    \mathcal{L}(s,y;a,b) = ys - (1-y)\log((s-a)(b-s))
\end{equation}
according to Eq. \ref{eq:new_loss_formula}.

For machine learning tasks that utilize neural networks, the output of the final layer is typically passed through a non-linear activation function that is bounded. For example, neural networks with sigmoid activation functions in the final layer are commonly used in binary classification tasks. In this case, the model outputs are limited to the interval $(0,1)$ and the ratio trick transformation given by Eq. \ref{eq:REVERT_Ratio_Trick_General_Form} becomes:
\begin{equation}\label{eq:sigmoid_ratio_trick}
    T(s) = \frac{1}{s} + \frac{1}{s-1} = \frac{1-2s}{s(1-s)},
\end{equation}
with corresponding loss function
\begin{equation}
    \mathcal{L}_{\rm REVERT}(s,y) = ys - (1-y)\log(s(1-s)) = ys - (1-y)(\log s + \log(1-s))
\end{equation}
referred to as the \textbf{REV}erse \textbf{E}ngineered \textbf{R}atio \textbf{T}rick (REVERT) loss. Based on this construction, if the risk is minimized to obtain the optimal model:
\begin{equation}
    s^* = \argmin_s\mathbb{E}_{X,Y}\left[Ys(X) - (1-Y)(\log s(X) + \log(1-s(X)))\right]
\end{equation}
then it follows that
\begin{equation}\label{eq:optimal_sigmoid_ratio_trick}
    \frac{q(\mathbf{x}|Y=1)}{q(\mathbf{x}|Y=0)} = r^*(\mathbf{x}) = T(s^*(\mathbf{x})) = \frac{1}{s^*(\mathbf{x})} + \frac{1}{s^*(\mathbf{x})-1}
\end{equation}
In other words, if this loss is used to train a classifier then the outputs of the learned model will be one-to-one with the density ratio through the transformation given in Eq. \ref{eq:sigmoid_ratio_trick}. Therefore, the task of quasiprobabilistic density ratio estimation is reduced to training a classifier using the REVERT loss.

In the case that a neural network with a sigmoid output activation function is used to model $s$, there is a simple relationship between the logits of the classifier and the density ratio. Let $z^*(\mathbf{x})$ represent the logit associated to the optimal classifier such that $s^*(\mathbf{x}) = \sigma(z^*(\mathbf{x}))$ where $\sigma$ is the sigmoid function. Substituting this relationship into Eq. \ref{eq:optimal_sigmoid_ratio_trick} yields:
\begin{equation}
    r^*(\mathbf{x}) = T(s^*(\mathbf{x})) = -2\rm{sinh}(z^*(\mathbf{x}))
\end{equation}
which provides a simple relationship between the logits of the optimal classifier and the density ratio \footnote{Only neural networks with sigmoid outputs are considered in this work, but these techniques can easily be used with other activation function outputs, such as hyperbolic tangent, by making the appropriate adjustments to the loss function and ratio trick transformation above. See Appendix \ref{app:AdditionalLosses} for some examples.}.

\section{Application to Particle Physics: Standard Model Effective Field Theory in High Energy Physics}
\label{sec:QP-SMEFT}

In this section, the proposed new quasiprobabilistic density ratio estimation loss function, REVERT, is applied to a representative problem encountered at the Large Hadron Collider (LHC). Proton-proton collisions at the LHC are simulated, in which pairs of Higgs bosons produced via gluon-gluon fusion ($gghh$) at a center of mass energy of $13$~TeV. Two datasets are generated under different hypotheses; the Standard Model (SM) and an extended theory of the Standard Model using the Standard Model Effective Field Theory (SMEFT) \cite{Grzadkowski_2010} paradigm. The goal is to use the density ratio estimate as an importance sampling based re-weighting technique, that maps expected values under one distribution to another. 

\subsection{Data Generation}
Higgs boson pairs ($hh$) produced via gluon-gluon fusion ($gg$) from proton-proton collisions at a center of mass energy of $\sqrt{s} = 13$ TeV are simulated at Next-to-Leading order (NLO) in the strong coupling constant $\alpha_{S}$ (QCD). The Higgs bosons are both forced to decay to muons, forming a final state with four muons from resonant Higgs decays, and additional final state partons in the form of gluons and quarks. 

Two samples were generated, a reference sample representing the Standard Model $gg\to hh$ process at NLO in QCD, and a target sample in the form of a beyond the Standard Model (BSM) $gg\to hh$ process generated using the SMEFT basis at NLO in QCD~\cite{Grzadkowski_2010,Brivio_2019,BUCHMULLER1986621}. The parameter configurations of each sample are given by Table \ref{tab:ggHH_configs}, where for the BSM target the transition amplitude is truncated according to equation 2.7(a) of Ref. \cite{Heinrich_2022}.

Each event is described by sixteen features used in training:
\begin{itemize}
\item Highest transverse momentum jet four-vector: $(p_{T,j}, \eta_{j}, \phi_{j}, m_{j})$
\item Muon four-vector momentum (x4): $(p_{T,\mu}, \eta_{\mu}, \phi_{\mu})$,
\end{itemize}
where $p_{T}$, $\eta$, $\phi$, and $m$ are the transverse momentum, pseudo-rapidity, azimuthal angle, and mass of that object. The di-Higgs invariant mass ($m_{hh}$) can be reconstructed from these features but is not explicitly used as a feature in training. However, this feature is used to evaluate the performance of the density ratio estimation models. After filters were applied to the data, 65\% of  the samples were used for training, 15\% for validation, and 20\% for testing, which corresponds to sample sizes per class (reference and target) of 776,369 for training, 179,162 for validation, and 238,883 for testing. For more details on this dataset, see Ref. \cite{drnevich2024neuralquasiprobabilisticlikelihoodratio}.

\begin{table*}
\caption{Proton-Proton collision $gg\to hh$ samples at $13.0$ TeV configurations}
\label{tab:ggHH_configs}
\centering
\begin{tblr}{colspec={p{4.5cm}||c|c|c|c|c}}
 \hline
 \SetCell[c=1]{c}{\centering \textbf{Sample}} & \SetCell[c=5]{c}{\textbf{SMEFT Wilson Coefficients}} \\
 \hline
 & $C_{H,kin}$ & $C_{H}$ & $C_{uH}$ & $C_{HG}$ & $\Lambda$ [TeV] \\
 \hline
 Standard Model (Ref.) & 0.0 & 0.0 & 0.0 & 0.0 & 0.0  \\
 \hline
 BSM (Target) & 13.5 & 2.64 & 12.6 & 0.0387 & 1.0  \\
 \hline
\end{tblr}
\end{table*}

\subsection{Machine Learning Setup}

Three neural models are trained on this dataset using the REVERT loss function of Eq. \ref{eq:sigmoid_ratio_trick}: an MLP neural network, and two variants of the \textit{Ratio of Signed Mixtures Model} (RoSMM) introduced in Ref.~\cite{drnevich2024neuralquasiprobabilisticlikelihoodratio}:  $\textrm{RoSMM}_c$ and $\textrm{RoSMM}_r$. In the former, a single MLP learns directly the quasiprobabilistic density ratio of Eq. \ref{eq:QLR}. In the latter, the quasiprobabilistic densities appearing in Eq. \ref{eq:QLR} are decomposed into their positively and negatively weighted components. The resulting density ratio takes the form of a mixture model:
\begin{equation}\label{eq:rhat}
\begin{split}
    r^*(\mathbf{x};c) &= \frac{q(\mathbf{x}|Y=1)}{q(\mathbf{x}|Y=0)} \\
                          &= \frac{c p_{w_+}(\mathbf{x}|Y=1) + (1-c)p_{w_-}(\mathbf{x}|Y=1)}{p_{w_+}(\mathbf{x}|Y=0)} \\
                          &= c r^{*}_{++}(\mathbf{x}) + (1-c)r^{*}_{+-}(\mathbf{x}),
\end{split}
\end{equation}
where $c \in [1,\infty)$ is a scalar mixture coefficient, and:
\begin{equation}\label{eq:RoSMM_subratios}
    r^*_{++}(\mathbf{x}) \equiv \frac{p_{w_+}(\mathbf{x}|Y=1)}{p_{w_+}(\mathbf{x}|Y=0)}, \; \; r^*_{+-}(\mathbf{x}) \equiv \frac{p_{w_-}(\mathbf{x}|Y=1)}{p_{w_+}(\mathbf{x}|Y=0)},
\end{equation}
are the sub-density ratios corresponding to the probabilistic density ratios of the target BSM positively weighted ($w_{+} > 0$) and negatively weighted ($w_{-} < 0$) distributions to the reference SM distribution, respectively. 

The RoSMM models are therefore composed of two pre-trained sub-density ratio estimation models and a mixture coefficient. The $\textrm{RoSMM}_c$ variant model leaves the sub-density ratio models frozen while optimizing the mixture coefficient. Meanwhile, the $\textrm{RoSMM}_r$ variant model allows the sub-density ratios and coefficients to be optimized simultaneously \footnote{See Ref.~\cite{drnevich2024neuralquasiprobabilisticlikelihoodratio} for more details}. In this work, the $\textrm{RoSMM}_c$ and $\textrm{RoSMM}_r$ models use the same pre-trained sub-density ratio models as in Ref.~\cite{drnevich2024neuralquasiprobabilisticlikelihoodratio} but use the new $\mathcal{L}_{\textrm{REVERT}}$ loss for the final optimization step.

All models were constructed and trained using PyTorch with the Adam optimizer \cite{NEURIPS2019_9015}. Training is stopped after a certain number (labeled ``stopping patience'') of sequential epochs where the validation loss is greater than the lowest validation loss across all epochs. An epoch is defined as either the entire training dataset or $10^5$ training samples, whichever is smaller. The number of parameters in the models were chosen such that each model had roughly the same number of total parameters and expressibility. The neural network architectures and optimization hyperparameter settings are summarized in Tables \ref{tab:architectures} \& \ref{tab:hyperparams}.

\begin{table}[ht]
    \centering
    \caption{Comparison of neural network architectures}
    \begin{tabular}{>{\raggedright\arraybackslash}p{3cm} 
                    >{\raggedright\arraybackslash}p{4cm} 
                    >{\raggedright\arraybackslash}p{4cm}}
        \toprule
        \textbf{Layer} & \textbf{MLP} & \textbf{$r_{++}$ and $r_{+-}$} \\
        \midrule
        Input          & $16$                  & $16$               \\
        Hidden 1       & Dense(128), ReLU      & Dense(128), ReLU   \\
        Hidden 2       & Dense(256), ReLU      & Dense(128), ReLU   \\
        Hidden 3       & Dense(128), ReLU      & Dense(128), ReLU   \\
        Output         & Dense(1), Sigmoid     & Dense(1), Sigmoid \\
        \bottomrule
    \end{tabular}
    \label{tab:architectures}
\end{table}

\begin{table}[ht]
    \centering
    \caption{Hyperparameter configurations for each model}
    \begin{tabular}{>{\raggedright\arraybackslash}p{4.5cm} 
                    >{\centering\arraybackslash}p{2cm} 
                    >{\centering\arraybackslash}p{2cm}
                    >{\centering\arraybackslash}p{2cm}
                    >{\centering\arraybackslash}p{2.5cm}}
        \toprule
        \textbf{Hyperparameter} & \textbf{MLP} & \textbf{$r_{++}$} & \textbf{$r_{+-}$} & \textbf{$\textrm{RoSMM}_{c(r)}$}\\
        \midrule
        Loss Function           & REVERT                & BCE          & BCE       & REVERT \\
        Learning rate           & $3\times 10^{-4}$     & $10^{-3}$    & $10^{-3}$ & $3\times 10^{-4}$    \\
        Batch size              & 256                   & 128          & 64        & 512  \\
        Optimizer               & Adam                  & Adam         & Adam      & Adam \\
        Stopping patience & 20                    & 15           & 15        & 10    \\
        \bottomrule
    \end{tabular}
    \label{tab:hyperparams}
\end{table}

\subsection{Performance Measures}

In order to evaluate the effectiveness of each model, the approximate density ratio is used as an importance weight to reweight a reference sample (Standard Model) to a target sample (BSM). A performance measure is then used to quantify the similarity, or agreement, between the reweighted reference sample and the target sample. To provide a direct comparison with the results from Ref.~\cite{drnevich2024neuralquasiprobabilisticlikelihoodratio}, the $\chi^2$ statistic and Tsallis relative entropy are used to quantify the similarity between one-dimensional histogram representations of the data\footnote{Details are available in Appendix A.5 of Ref.~\cite{drnevich2024neuralquasiprobabilisticlikelihoodratio}}. However, these are challenging to apply in high-dimensional settings as they require multi-dimensional histograms or density estimation. To supplement these, a new multivariate distance is introduced which provides a summary measure of the performance of each model. This multivariate distance is an extended form of the Sliced-Wasserstein distance that is compatible with quasiprobability distributions.

\subsubsection{Extended Sliced-Wasserstein Distance}

Extensions of the $p$-Wasserstein distance \footnote{See Appendix \ref{app:SWD} for a review of Wasserstein metrics} to signed measures was explored by Ambrosio et al. \cite{AMBROSIO2011217} where they suggested the following extension. If $\mu,\nu$ are signed Radon measures on $\mathbb{R}^d$ with finite mass, then the extended $1$-Wasserstein distance is:
\begin{equation}
    \mathbb{W}_1(\mu, \nu) = \mathbf{W}_1(\mu^+ + \nu^-, \nu^+ + \mu^-)
\end{equation}
where $\mu = \mu^+ - \mu^-$ and $\nu = \nu^+ - \nu^-$ are the Hahn-Jordan decompositions \footnote{This extension can be applied to any $p$-Wasserstein distance, but it is only a distance for $p=1$}. Additionally, through the same arguments given by Piccoli et al. in Proposition 17 of Ref. \cite{piccoli2019wassersteinnormsignedmeasures}, one can show that $\mathbb{W}_1(\mu, \nu)$ can be computed with any decomposition of the signed measures - not just the Hahn-Jordan decomposition. This motivates our following definition of the extended Sliced 1-Wasserstein distance for quasiprobability distributions.

\bdfn
Let $\mu,\nu$ be signed probability measures on $\mathbb{R}^d$ and $\mu^+, \mu^-, \nu^+, \nu^-$  be Radon measures on $\mathbb{R}^d$ with finite mass such that $\mu = \mu^+ - \mu^-$ and $\nu = \nu^+ - \nu^-$. We define the \emph{extended Sliced 1-Wasserstein distance} between $\mu$ and $\nu$ to be
\begin{equation}
    \mathbb{SW}_1(\mu, \nu) = \int_{\mathbb{S}^{d-1}}\mathbb{W}_1(\pi^\theta_\#\mu, \pi^\theta_\#\nu)d\boldsymbol{\sigma}(\theta) = \mathbf{SW}_1(\mu^+ + \nu^-, \nu^+ + \mu^-)
\end{equation}
where $\boldsymbol{\sigma}$ denotes the uniform distribution on the $d-1$ sphere, $\pi^\theta$ denotes the linear form given by $\pi^\theta(x) \equiv\langle\theta, x\rangle$, and $\pi^\theta_\#$ denotes the push-forward operator associated to $\pi^\theta$.
\edfn
The fact that this is also a metric follows from the fact that $\mathbb{W}_1$ is a metric and integration over projections preserves metric properties \cite{10.1007/978-3-642-24785-9_37}. In the case where the measures $\mu$ and $\nu$ are nonnegative this reduces to the regular Sliced 1-Wasserstein distance. Notably, this is only a valid distance for $p=1$, which prevents the use of various efficient techniques developed for computing Sliced 2-Wasserstein distances. For simplicity, we will refer to the extended Sliced 1-Wasserstein distance as the Sliced-Wasserstein distance.

In order to apply this metric, a decomposition of the two quasiprobability distributions into positive and negative parts is needed. We choose the decomposition inspired by Ref. \cite{drnevich2024neuralquasiprobabilisticlikelihoodratio}, where $\mu^+$ and $\mu^-$ correspond to $\mu$ restricted to the positively and negatively weighted subsets, respectively. To provide a common scale for comparison, independent of sample size, the weights for the datasets representing $\mu^+ + \nu^-$ and $\nu^+ + \mu^-$ are normalized to sum to unity (see Algorithm \ref{alg:swd} for more details). After constructing the two appropriate sets, the Sliced-Wasserstein distance is computed using the \verb|sliced_wasserstein_distance| function from the Python Optimal Transport library \cite{flamary2021pot}.

\subsection{Results}

\begin{figure}[!t]
\centering
\includegraphics[scale=0.45]{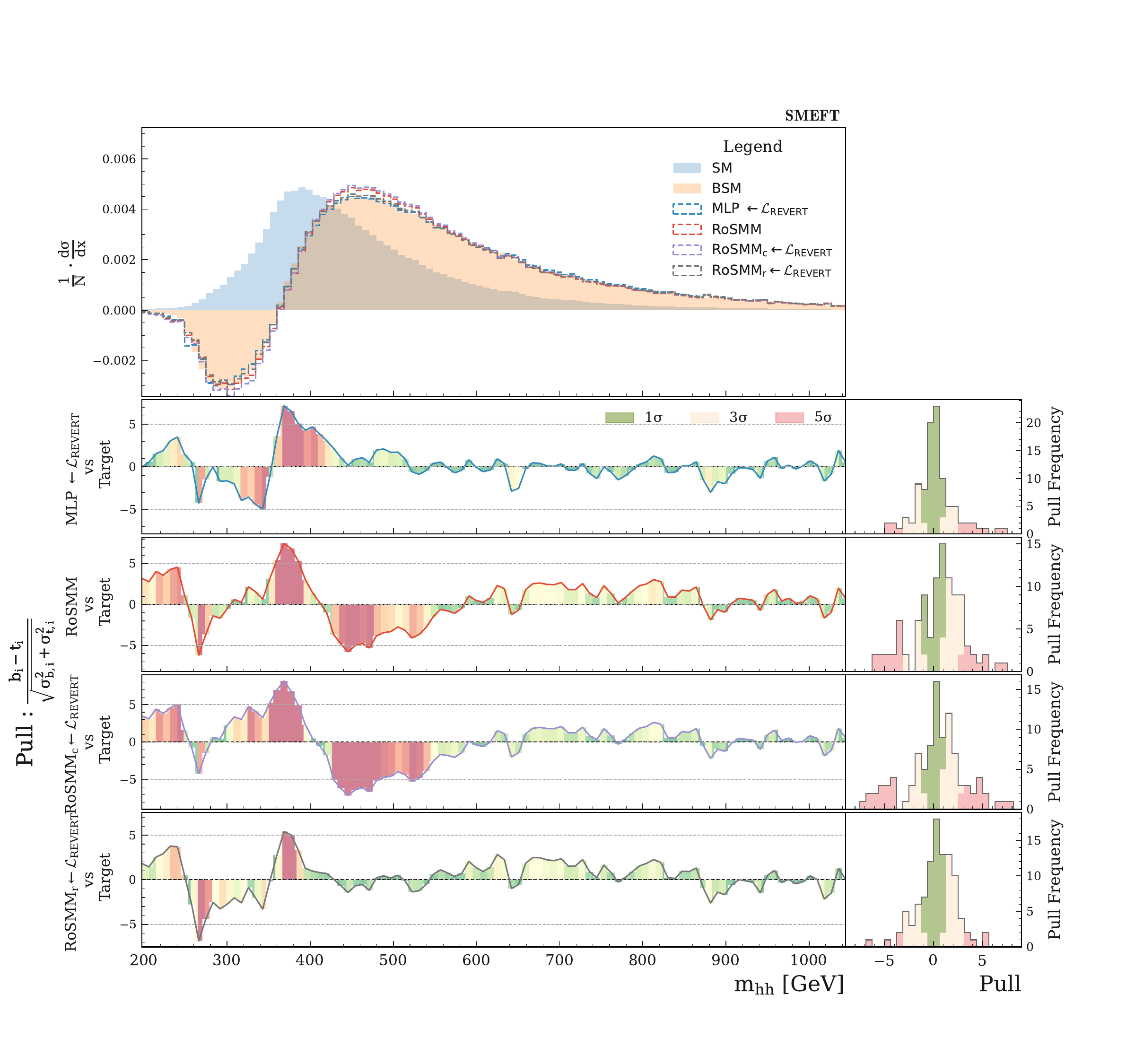}
\caption{The reweighting closure plots for the di-Higgs invariant mass ($m_{hh}$) where the reference (Standard Model) distribution is mapped to the target (SMEFT) distribution using the different density ratio estimation models.}\label{fig:smeft_mhh_reweighting}
\end{figure}

The holdout testing datasets are used to evaluate each of the models from Ref.\cite{drnevich2024neuralquasiprobabilisticlikelihoodratio} in addition to the MLP, $\textrm{RoSMM}_{c}$, and $\textrm{RoSMM}_{r}$ models trained using the new loss. For all measures, the BSM (target) dataset is compared to the Standard Model (reference) dataset weighted by the density ratios output by the corresponding model. The results for the extended form of the Sliced-Wasserstein distances are shown in Table \ref{tab:smeft_swd_results}. For each model, it is computed 1000 times using 50 projections each time. The results shown in Table \ref{tab:smeft_swd_results} report the mean plus/minus the standard deviation of those 1000 calculations.

The results of each distance measure from Ref. \cite{drnevich2024neuralquasiprobabilisticlikelihoodratio} for each type of model and feature are shown in Table \ref{tab:smeft_x2_results} and Table \ref{tab:smeft_tsallis_results}. Figure \ref{fig:smeft_mhh_reweighting} illustrates the closure performance of the density ratio estimation models trained using the REVERT loss for the mass of the two Higgs system using the approximate density ratio as a weight that maps the reference sample to the target. Additional closure plots for each of the other features are shown in Appendix \ref{app:plots}.

These measures demonstrate that each model is capable of learning quasiprobabilistic density ratios. However, the MLP and the fully optimized ratio of signed mixtures model ($\textrm{RoSMM}_r$) trained using the REVERT loss performed the best overall according to all measures. This can be verified visually in Figure \ref{fig:smeft_mhh_reweighting} where all other models considerably overestimate the target density in the peak around 450--500 GeV. For this dataset, there is also no significant difference in performance between a standard MLP and a $\textrm{RoSMM}_r$ model when both are trained using the new loss, although the $\textrm{RoSMM}_r$ model indicates slightly better performance with the Sliced-Wasserstein distance.

\begin{table*}[!t]
\caption{SMEFT Results - $\mathbb{SW}_1$ Distance} \label{tab:smeft_swd_results}
\centering
\begin{tblr}{colspec={p{4.5cm}||c}}
 \hline
\centering \textbf{DRE Model} & \centering \textbf{Sliced-Wasserstein Distance: $\mu \pm \sigma$} \\
 \hline \hline
 MLP $\leftarrow \mathcal{L}_{\textrm{REVERT}}$ & $0.430 \pm 0.039$ \\
 \hline
 RoSMM & $0.810 \pm 0.060$ \\
 \hline
 $\textrm{RoSMM}_c \leftarrow \mathcal{L}_{\textrm{REVERT}}$ &  $0.902 \pm 0.071$ \\
 \hline
 $\textrm{RoSMM}_r \leftarrow \mathcal{L}_{\textrm{REVERT}}$ &  $0.301 \pm 0.022$ \\
 \hline
Reference &  $13.450 \pm 1.380$ \\
 \hline
\end{tblr}
\end{table*}

\begin{table*}[!t]
\caption{SMEFT Results - $\chi^2$ Scores} \label{tab:smeft_x2_results}
\centering
\begin{adjustbox}{width=0.99\textwidth}
\begin{tblr}{colspec={p{4.25cm}||c|c|c|c|c|c|c|c},stretch=0.5}
 \hline
 \SetCell[r=2]{m}{\centering \textbf{DRE Model}} & \SetCell[c=8]{c} \textbf{Feature} &&&&&&&\\
 \hline
 & 1st Muon $p_T$ & 2nd Muon $p_T$ & 3rd Muon $p_T$ & 4th Muon $p_T$ & Jet $p_T$ & Jet $m$ & $hh$ $p_T$ & $m_{hh}$ \\
 \hline \hline
 MLP $\leftarrow \mathcal{L}_{\textrm{REVERT}}$ & \textbf{1.03} & \textbf{1.05} & \textbf{0.87} & \textbf{1.08} & 1.93 & 1.00 & 2.36 & 2.78 \\
 \hline
 RoSMM & 1.78 & 1.76 & 1.10 & 1.23 & 2.20 & 1.65 & 2.08 & 4.45 \\
 \hline
 $\textrm{RoSMM}_c \leftarrow \mathcal{L}_{\textrm{REVERT}}$ & 2.08 & 2.38 & 1.26 & 1.15 & 2.37 & 1.61 & 2.10 & 5.52 \\
 \hline
 $\textrm{RoSMM}_r \leftarrow \mathcal{L}_{\textrm{REVERT}}$ & 1.16 & 1.19 & 1.00 & 1.27 & \textbf{1.55} & \textbf{0.96} & 2.12 & \textbf{2.65} \\
 \hline
\end{tblr}
\end{adjustbox}
\end{table*}

\begin{table*}[!t]
\caption{SMEFT Results - $D_{S_{2}}$ Values} \label{tab:smeft_tsallis_results}
\centering
\begin{adjustbox}{width=0.99\textwidth}
    \begin{tblr}{colspec={p{4.25cm}||c|c|c|c|c|c|c|c},stretch=0.5}
    \hline
    \SetCell[r=2]{m}{\centering \textbf{DRE Model}} & \SetCell[c=8]{c} \textbf{Feature} &&&&&&&\\
    \hline
    & 1st Muon $p_T$ & 2nd Muon $p_T$ & 3rd Muon $p_T$ & 4th Muon $p_T$ & Jet $p_T$ & Jet $m$ & $hh$ $p_T$ & $m_{hh}$ \\
    \hline \hline
    MLP $\leftarrow \mathcal{L}_{\textrm{REVERT}}$ & 0.0031 & 0.0052 & 0.0030 & \textbf{0.0033} & \textbf{0.0019} & 0.0051 & 0.012 & \textbf{0.0001} \\
    \hline
    RoSMM & 0.0022 & 0.0083 & 0.0032 & 0.0036 & 0.205 & 0.0071 & \textbf{0.0042} & 0.0046 \\
    \hline
    $\textrm{RoSMM}_c \leftarrow \mathcal{L}_{\textrm{REVERT}}$ & 0.033 & 0.19 & 0.0042 & 0.0036 & 0.017 & 0.0075 & 0.0062 & 0.024 \\
    \hline
    $\textrm{RoSMM}_r \leftarrow \mathcal{L}_{\textrm{REVERT}}$ & \textbf{0.0014} & \textbf{0.0021} & \textbf{0.0029} & 0.0035 & 0.0048 & \textbf{0.0034} & 0.016 & 0.0003 \\
    \hline
    \end{tblr}
\end{adjustbox}
\end{table*}

\section{Conclusions}
\label{sec:Conclusions}
In this work, we considered a generalization of the classifier-based density-ratio estimation task to a quasiprobabilistic setting where probability densities can be negative. Existing loss functions designed for this task implicitly define a relationship between the optimal classifier and the target quasiprobabilistic density ratio (referred to as a ``ratio trick'') which is discontinuous or not surjective. To address these problems, we demonstrate a general process for creating convex loss functions from a desired ``ratio trick'' transformation and use this to construct the REVERT loss function. The REVERT loss is convex and has a ``ratio trick'' transformation that is continuous and capable of transforming the optimal classifier outputs to any real-valued density ratio. Hence, training a classifier with this loss enables quasiprobabilistic density ratio estimation through the associated ``ratio trick'' transformation. This loss function also serves as a substitute for the PARE loss given in Ref.~\cite{drnevich2024neuralquasiprobabilisticlikelihoodratio} when optimizing Ratio of Signed Mixtures Models (RoSMMs) since it circumvents the hyper-parameter optimisation problem of the pole adjustable hyper-parameters $t_{0/1}$ in the PARE loss. The effectiveness of these techniques were demonstrated with a realistic particle physics example with a simple neural network classifier as well as Ratio of Signed Mixtures Models (RoSMMs) to directly compare with the results from Ref.~\cite{drnevich2024neuralquasiprobabilisticlikelihoodratio}. To quantify performance, an extended form of the Sliced-Wasserstein distance was introduced to provide a metric between multivariate quasiprobability distributions. Based on all performance measures, models trained using the REVERT loss function, for classification-based quasiprobabilistic density ratio estimation, achieved state-of-the-art performance.

\section*{References}
\printbibliography[heading=none, title={References}]

\section*{Appendix}
\label{Appendix}

\subsection{Notation}
\label{app:Notation}
Some commonly used notation is listed here to avoid any confusion:
\begin{align*}
    \mathbb{E}_{X}\left[f(X)\right] &= \int f(x)p(x)dx\\
    \mathbb{E}_{X,Y}\left[f(X,Y)\right] &= \begin{cases}
        \int\int f(x,y)p(x,y)dxdy & \text{if Y is continuous} \\
        \int\sum_{y} f(x,y)p(x,y)dx & \text{if Y is discrete}
    \end{cases}\\
    \mathbb{E}_{X|Y=y}\left[f(X,Y)\right] &= \mathbb{E}\left[f(X,Y)|Y=y\right] = \int f(x,y)p(x|y)dx
\end{align*}
where $p(\cdot)$ is the probability density function and $f$ is an arbitrary function.

\subsection{Risk Extrema}
\label{app:LossDerivation}
Here we demonstrate how to solve for the extrema of the risk for a loss function given by Eq. \ref{eq:classification_loss}. Consider the risk defined as follows:

\begin{align*}
    R[s] &= \mathbb{E}_{X,Y}\left[\mathcal{L}(s(X),Y)\right] \\
    &= \mathbb{E}_{X}\mathbb{E}_{Y|X}\left[\mathcal{L}(s(X),Y)\right] \\
    &= \int_{\mathcal{X}}\mathbb{E}_{Y|X=\mathbf{x}}\left[\mathcal{L}(s(X),Y)\right]q(\mathbf{x})d\mathbf{x}
\end{align*}
We define the Lagrangian as the integrand of the previous equation:
\begin{equation*}
    L(\mathbf{x},s) = \mathbb{E}_{Y|X=\mathbf{x}}\left[\mathcal{L}(s(X),Y)\right]q(\mathbf{x})
\end{equation*}
such that
\begin{equation*}
    R[s] = \int_{\mathcal{X}}L(\mathbf{x},s)d\mathbf{x} ~.
\end{equation*}
The extrema of $R$ can be found by using the Euler-Lagrange equation, which says that if $s^*$ is an extremum of $R$ then:
\begin{align*}
    \left. \frac{\partial L}{\partial s}\right|_{s=s^*(\mathbf{x})} = 0
\end{align*}
Before solving this equation, we start by simplifying the Lagrangian using properties of expectations:
\begin{align*}
    L(\mathbf{x},s) &= \mathbb{E}_{Y|X=\mathbf{x}}\left[\mathcal{L}(s,Y)\right]q(\mathbf{x}) \\
    &= \mathbb{E}_{Y|X=\mathbf{x}}\left[Yf(s) + (1-Y)g(s)\right]q(\mathbf{x}) \\
    &= \left(\mathbb{E}_{Y|X=\mathbf{x}}\left[Y\right]f(s) + \mathbb{E}_{Y|X=\mathbf{x}}\left[1-Y\right]g(s)\right)q(\mathbf{x}) \\
    &= \left(p(Y=1|\mathbf{x})f(s) + p(Y=0|\mathbf{x})g(s)\right)q(\mathbf{x})
\end{align*}
Then we can compute the solution to the Euler-Lagrange equation:
\begin{align*}
    \left. \frac{\partial L}{\partial s}\right|_{s=s^*(\mathbf{x})} &= 0 \\
    \implies \frac{\partial}{\partial s}\left[\left(p(Y=1|\mathbf{x})f(s) + p(Y=0|\mathbf{x})g(s)\right)q(\mathbf{x})\right]_{s=s^*(\mathbf{x})} &= 0 \\
    \implies \left(p(Y=1|\mathbf{x})f'(s^*(\mathbf{x})) + p(Y=0|\mathbf{x})g'(s^*(\mathbf{x}))\right)q(\mathbf{x}) &= 0 \\
    \implies p(Y=1|\mathbf{x})f'(s^*(\mathbf{x})) + p(Y=0|\mathbf{x})g'(s^*(\mathbf{x})) &= 0 \\
    \implies \frac{p(Y=1|\mathbf{x})}{p(Y=0|\mathbf{x})} &= -\frac{g'(s^*(\mathbf{x}))}{f'(s^*(\mathbf{x}))}
\end{align*}
as long as $q(\mathbf{x}) \neq 0$. Additionally, since $P(y|\mathbf{x}) = q(\mathbf{x}|y)P(y)/q(\mathbf{x})$, in the case where the training datasets are equal sizes we have $p(Y=0)=p(Y=1)=1/2$ and consequently
\begin{align*}
    -\frac{g'(s^*(\mathbf{x}))}{f'(s^*(\mathbf{x}))} = \frac{q(\mathbf{x}|Y=1)}{q(\mathbf{x}|Y=0)} = r^*(\mathbf{x})
\end{align*}

\subsection{Convexity of the Loss}
\label{app:Convexity}
For the derivations presented here, we adopt the same notation, constructions, and assumptions introduced in Section~\ref{sec:Losses}. To simplify notation, we define the following sets. Let $B = (a,b) \subseteq \mathbb{R}$ for some $a,b \in \mathbb{R}$ with $a<b$, and let $C^2(\bar{\mathcal{X}},B)$ denote the set of continuous, twice-differentiable functions mapping from the closure of $\mathcal{X}$ to $B$.

\bprop
Let $T:B \to \mathbb{R}$ be a homeomorphism. The Lagrangian
\begin{equation}
    L(\mathbf{x},s) = \mathbb{E}_{Y|X=\mathbf{x}}\!\left[\mathcal{L}(s,Y)\right]q(\mathbf{x}),
\end{equation}
with
\begin{equation}
    \mathcal{L}(s,y) = ys - (1-y)\int T(s)\,ds,
\end{equation}
is convex in $s$, up to the redefinition $T \mapsto -T$, if and only if $q(\mathbf{x}| Y=0)$ is nonnegative.
\eprop

\bpf
We begin with preliminary observations. Define $g(s) = \int -T(s)\,ds$. Then $g'(s) = -T(s)$, so $g$ is convex whenever $-T$ is monotonically increasing. Since $T$ is a homeomorphism on $\mathbb{R}$, it must be strictly monotone. If $T$ is strictly decreasing, then $-T$ is strictly increasing, and $g$ is convex. If $T$ is strictly increasing, we may redefine $T \mapsto -T$, which remains a homeomorphism from $(a,b)$ to $\mathbb{R}$, ensuring convexity of $g(s)$. Without loss of generality, we assume $T$ has been redefined if necessary so that $g(s)$ is convex.

Next, consider the Lagrangian from Eq. \ref{eq:lagrangian} when the training dataset is equally balanced such that $p(Y=0)=p(Y=1)=1/2$:
\begin{align*}
    L(\mathbf{x},s) &= q(\mathbf{x}|Y=1)p(Y=1)s + q(\mathbf{x}|Y=0)p(Y=0)g(s) \\
    &= \tfrac{1}{2}\big(q(\mathbf{x}|Y=1)s + q(\mathbf{x}|Y=0)g(s)\big).
\end{align*}
This representation will be used to prove both directions.

\paragraph{Forward direction (convexity $\implies q(\mathbf{x}|Y=0)\geq 0$).}  
Suppose, for contradiction, that there exists $\mathbf{x}_0 \in \mathcal{X}$ with $q(\mathbf{x}_0|Y=0)<0$. Then the mapping $s \mapsto q(\mathbf{x}_0|Y=0)g(s)$ is concave and for any $s_1,s_2 \in C^2(\bar{\mathcal{X}},B)$ and $t \in [0,1]$,
\begin{align*}
    L(\mathbf{x}_0,ts_1 + (1-t)s_2) &= \frac{1}{2}\left(q(\mathbf{x}_0|Y=1)(ts_1 + (1-t)s_2) + q(\mathbf{x}_0|Y=0)g(ts_1 + (1-t)s_2)\right) \\
    &\geq \frac{1}{2}\left(q(\mathbf{x}_0|Y=1)(ts_1 + (1-t)s_2) + q(\mathbf{x}_0|Y=0)(tg(s_1) + (1-t)g(s_2))\right) \\
    &= t~  \frac{1}{2}\left(q(\mathbf{x}_0|Y=1)s_1 + q(\mathbf{x}_0|Y=0)g(s_1)\right) \\
    &\quad + (1-t)~\frac{1}{2}\left(q(\mathbf{x}_0|Y=1)s_2 + q(\mathbf{x}_0|Y=0)g(s_2)\right) \\
    &= t~L(\mathbf{x}_0,s_1) + (1-t)~L(\mathbf{x}_0,s_2)
\end{align*}
Thus $L$ is concave in $s$ at $\mathbf{x}_0$, contradicting convexity. Hence $q(\mathbf{x}| Y=0)$ must be nonnegative.

\paragraph{Reverse direction ($q(\mathbf{x}|Y=0)\geq 0 \implies$ convexity).}
If $q(\mathbf{x}| Y=0)\geq 0$, then $s \mapsto q(\mathbf{x}| Y=0)g(s)$ is convex. Through the same steps shown for the previous direction, for any $s_1,s_2 \in C^2(\bar{\mathcal{X}},B)$ and $t \in [0,1]$,
\begin{align*}
    L(\mathbf{x},ts_1+(1-t)s_2) 
    &\leq t\,L(\mathbf{x},s_1) + (1-t)\,L(\mathbf{x},s_2).
\end{align*}
Thus $L$ is convex in $s$.
\epf

\bcor
If $q(\mathbf{x}|Y=0)$ is nonnegative, then the minimizer of the risk given by
\begin{equation}
    \mathbb{E}_{X,Y}\left[\mathcal{L}(s(X),Y)\right]
\end{equation}
exists and is the unique function $s^*$ defined as
\begin{equation}
    s^*(\mathbf{x}) = T^{-1}(r^*(\mathbf{x}))
\end{equation}
up to a redefinition $T\mapsto -T$.
\ecor

\bpf
This fact follows from rewriting the risk as follows
\begin{equation*}
    \mathbb{E}_{X,Y}\left[\mathcal{L}(s(X),Y)\right] = \int_{\mathcal{X}}\mathbb{E}_{Y|X=\mathbf{x}}\left[\mathcal{L}(s(\mathbf{x}),Y)\right]q(\mathbf{x})d\mathbf{x} = \int_{\mathcal{X}} L(\mathbf{x}, s(\mathbf{x}))d\mathbf{x}
\end{equation*}
and using the previous proposition to conclude that the integrand is convex. This ensures that the solution to the Euler-Lagrange equation is the function that minimizes this loss \cite{alma990009844240107876}. Uniqueness comes from the fact that $T$ is injective and that any function which extremizes the risk must satisfy $s^*(\mathbf{x}) = T^{-1}(r^*(\mathbf{x}))$.
\epf

\subsection{Additional Loss Functions}
\label{app:AdditionalLosses}
There are many potential choices for the homeomorphism $T:(a,b) \to \mathbb{R}$. Although this paper focuses on a particular one, we list a few other ``ratio tricks'' and their corresponding loss functions in Table \ref{tab:additional_homeos} as a reference.

\begin{table*}
\caption{Example ``Ratio Trick'' Homeomorphisms and Loss Functions}
\label{tab:additional_homeos}
\centering
\begin{adjustbox}{width=0.99\textwidth}
\begin{tblr}{colspec={c|c|c}}
 \hline
 
 \SetCell[c=1]{c}{\centering \textbf{Codomain of} $s$} & \SetCell[c=1]{c}{\textbf{Homeomorphism}} $T(s)$ & \SetCell[c=1]{c}{\textbf{Loss Function} $\mathcal{L}(s,y)$}\\
 \hline
 $(0,1)$ & $\ln(1-s) - \ln(s)$ & $ys + (1-y)\left(s\ln\left(s\right) + \left(1-s\right)\ln\left(1-s\right)\right)$ \\
 \hline
 $(0,1)$ & $-\tan(\pi(s-1/2))$ & $ys-\left(1-y\right)\frac{1}{\pi}\ln\left(\sin\left(\pi s\right)\right)$   \\
 \hline
 $(0,1)$ & $\begin{cases}
      -\ln(2s) & 0 < s \leq 0.5 \\
      \ln(2-2s) & 0.5 \leq s < 1
  \end{cases}$ & $ys + (1-y)\begin{cases}
      s\left(\ln\left(2s\right)-1\right)+1 & 0 < s \leq 0.5 \\
      \left(1-s\right)\ln\left(2-2s\right)+s & 0.5 \leq s < 1
  \end{cases}$ \\
  \hline
  $(-1,1)$ & $s / (|s|-1)$ & $ys - (1-y)\frac{1}{2}\left(\ln(1-s^2) +2|s| - \ln\left(\frac{1 + s}{1-s}\right) \textrm{sgn}(s)\right)$ \\
 \hline 
 $(-1,1)$ & $1/(s+1) + 1/(s-1)$ & $ys - (1-y)(\log(s+1) + \log(1-s))$ \\
 \hline
\end{tblr}
\end{adjustbox}
\end{table*}

\subsection{Background on Wasserstein Distances}
\label{app:SWD}
There has been extensive work on defining suitable metrics between probability distributions, but many can be too computationally expensive to estimate using only samples from the distributions, particularly in high dimensions. One approach to alleviate this issue was the introduction of the Sliced-Wasserstein distance (SW) \cite{10.1007/978-3-642-24785-9_37}. Rather than computing the Wasserstein distance in $d$-dimensions, the Sliced-Wasserstein distance requires computing many one-dimensional projections of the data and then computing the one-dimensional Wasserstein distance for each projection. This alone makes the computation much more feasible for high-dimensional data, but further work demonstrated that the Sliced-Wasserstein distance can be approximated efficiently under certain settings \cite{nadjahi2021fast}. 

Some important definitions and background are provided as a review, but we refer the reader to the appropriate literature for the precise mathematical requirements and additional details not mentioned here. In particular, for results in the probabilistic setting Ref. \cite{nadjahi:tel-03533097} provides a comprehensive review. First, we introduce the $p$-Wasserstein distance which is defined as follows:

\bdfn
Let $p \in [1,\infty)$ and $\mu,\nu$ be probability measures on $\mathbb{R}^d$ with finite $p$-th moment. The \emph{p-Wasserstein distance} between $\mu$ and $\nu$ is given by
\begin{equation}
    \mathbf{W}_p(\mu, \nu) = \left(\inf_{\gamma \in \Gamma(\mu,\nu)}\int_{\mathbb{R}^d\times\mathbb{R}^d} \norm{x-y}^p d\gamma(x,y)\right)^{1/p}
\end{equation}
where $\norm{\cdot}$ is the usual Euclidean norm and $\Pi(\mu,\nu)$ is the set of joint probability measures on $\mathbb{R}^d\times\mathbb{R}^d$ with marginals in the first and second variables given by $\mu$ and $\nu$, respectively.
\edfn
In general, this computation has super-cubic cost to compute, making it infeasible for many applications \cite{peyré2020computationaloptimaltransport}. However, for one-dimensional distributions this effectively reduces to a sorting problem, which is relatively efficient to compute. The Sliced-Wasserstein distance leverages this fact to provide a more computationally efficient alternative to $\mathbf{W}_p$. Using the same definitions, the \emph{Sliced p-Wasserstein distance} \cite{10.1007/978-3-642-24785-9_37} is given by
\begin{equation}
    \mathbf{SW}_p(\mu, \nu) = \left(\int_{\mathbb{S}^{d-1}}\mathbf{W}^p_p(\pi^\theta_\#\mu, \pi^\theta_\#\nu)d\boldsymbol{\sigma}(\theta)\right)^{1/p}
\end{equation}
where $\boldsymbol{\sigma}$ denotes the uniform distribution on the $d-1$ sphere, $\pi^\theta$ denotes the linear form given by $\pi^\theta(x) \equiv\langle\theta, x\rangle$, and $\pi^\theta_\#$ denotes the push-forward operator associated to $\pi^\theta$. Hence the computation involves taking the expectation of one-dimensional Wasserstein distances uniformly over the $d-1$ sphere. If $L$ samples are drawn from the $d-1$ sphere to estimate the expectation using the Monte Carlo method, then this metric has computational complexity $\mathcal{O}(Ldn+Ln\log(n))$ \cite{nadjahi2021fast}. Further work has been done to estimate this metric with more computationally efficient methods, but they are restricted to the setting $p=2$ \cite{nadjahi2021fast}.

\begin{algorithm}
\caption{Computing the Extended Sliced 1-Wasserstein Distance}\label{alg:swd}
\begin{algorithmic}
\Require Arrays $\rm{X}_\mu, \rm{W}_\mu, \rm{X}_\nu, \rm{W}_\nu$ \Comment{Sample data and weights from $\mu$ and $\nu$}
\State $\rm{W}_\mu \leftarrow \{\rm{W}_\mu[i]~/~\rm{sum}(\rm{W}_\mu) : 1\leq i \leq \rm{size}(\rm{W}_\mu)\}$ \Comment{Normalize the weights for $\mu$}
\State $\rm{W}_\nu \leftarrow \{\rm{W}_\nu[i]~/~\rm{sum}(\rm{W}_\nu) : 1\leq i \leq \rm{size}(\rm{W}_\nu)\}$ \Comment{Normalize the weights for $\nu$}
\State $\rm{PosX}_\mu, \rm{NegX}_\mu, \rm{PosW}_\mu, \rm{NegW}_\mu \leftarrow \{\}, \{\}, \{\}, \{\}$
\For{$i = 1$ \textbf{to} size($\rm{X}_\mu$)}
    \If{$\rm{W}_\mu[i] \geq 0$}
        \State $\rm{PosX}_\mu \leftarrow \rm{PosX}_\mu \cup \{\rm{X}_\mu[i]\}$
        \State $\rm{PosW}_\mu \leftarrow \rm{PosW}_\mu \cup \{\rm{W}_\mu[i]\}$
    \Else
        \State $\rm{NegX}_\mu \leftarrow \rm{NegX}_\mu \cup \{\rm{X}_\mu[i]\}$
        \State $\rm{NegW}_\mu \leftarrow \rm{NegW}_\mu \cup \{\rm{W}_\mu[i]\}$
    \EndIf
\EndFor
\State Split $\rm{X}_\nu, \rm{W}_\nu$ into $\rm{PosX}_\nu, \rm{NegX}_\nu, \rm{PosW}_\nu, \rm{NegW}_\nu$ in the same way
\State $\rm{ComboX}_1, \rm{ComboW}_1 \leftarrow \rm{PosX}_\mu \cup \rm{NegX}_\nu,\ \rm{PosW}_\mu \cup \rm{NegW}_\nu$
\State $\rm{ComboX}_2, \rm{ComboW}_2 \leftarrow \rm{PosX}_\nu \cup \rm{NegX}_\mu,\ \rm{PosW}_\nu \cup \rm{NegW}_\mu$
\State $\rm{ComboW}_1 \leftarrow \{\rm{ComboW}_1[i]~/~\rm{sum}(\rm{ComboW}_1) : 1\leq i \leq \rm{size}(\rm{ComboW}_1)\}$
\State $\rm{ComboW}_2 \leftarrow \{\rm{ComboW}_2[i]~/~\rm{sum}(\rm{ComboW}_2) : 1\leq i \leq \rm{size}(\rm{ComboW}_2)\}$
\State \Return $\mathbf{SW}_1(\mu=\rm{ComboX}_1, \nu=\rm{ComboX}_2, W_\mu=\rm{ComboW}_1, W_\nu=\rm{ComboW}_2)$
\end{algorithmic}
\end{algorithm}

\subsection{Additional Plots}
\label{app:plots}
\begin{figure}[H]
\centering
\includegraphics[scale=0.21]{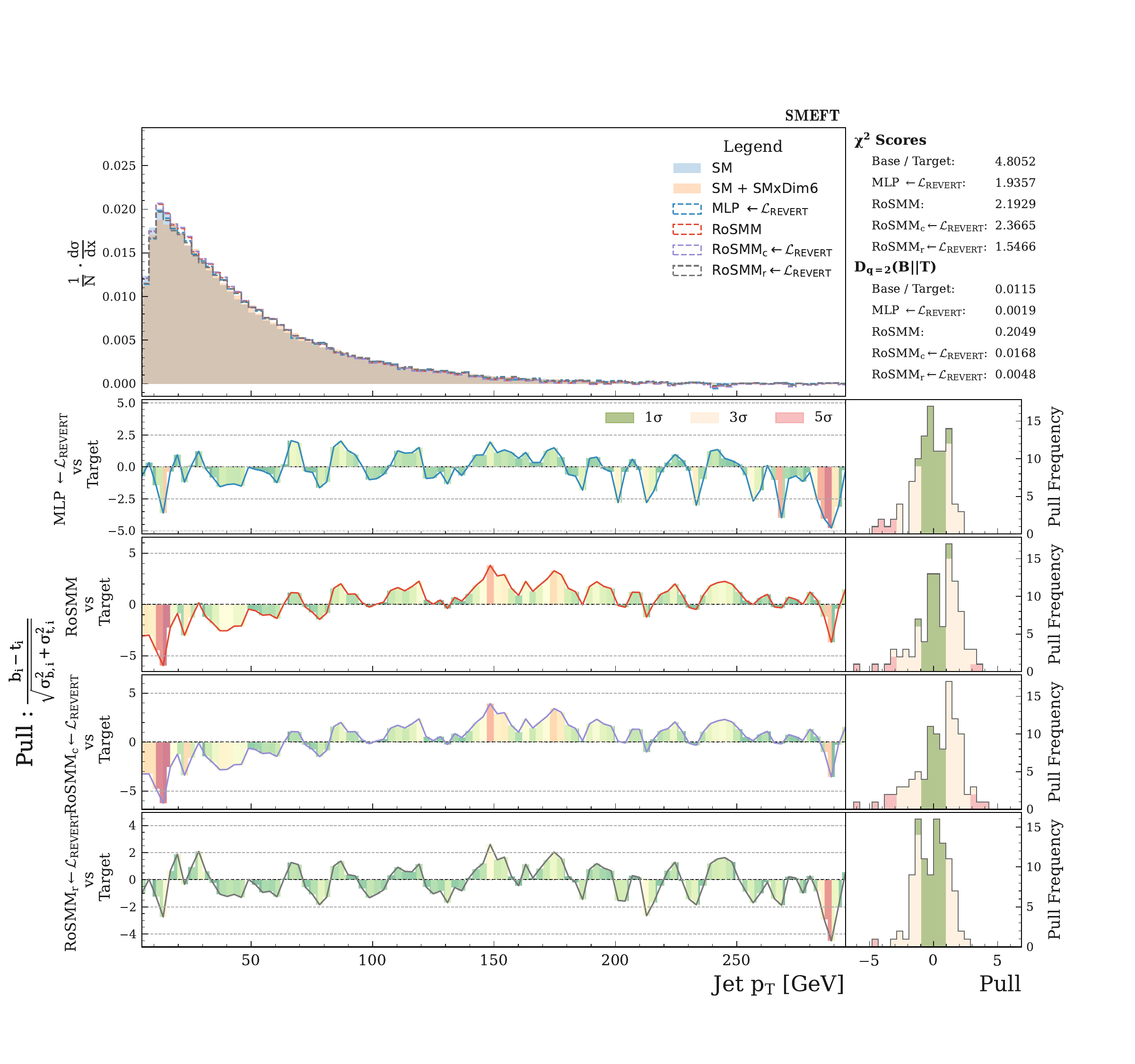}
\includegraphics[scale=0.21]{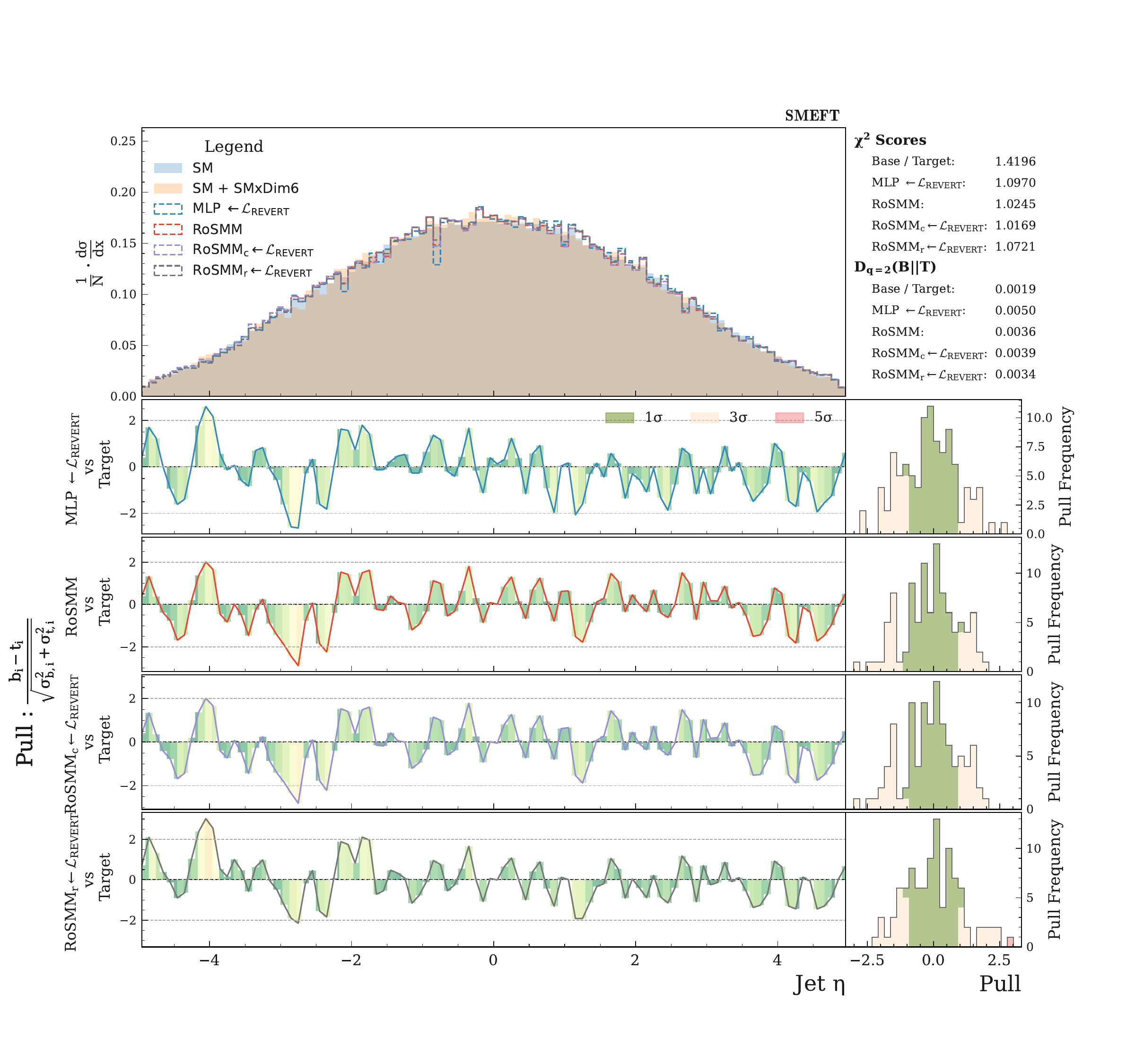}
\includegraphics[scale=0.21]{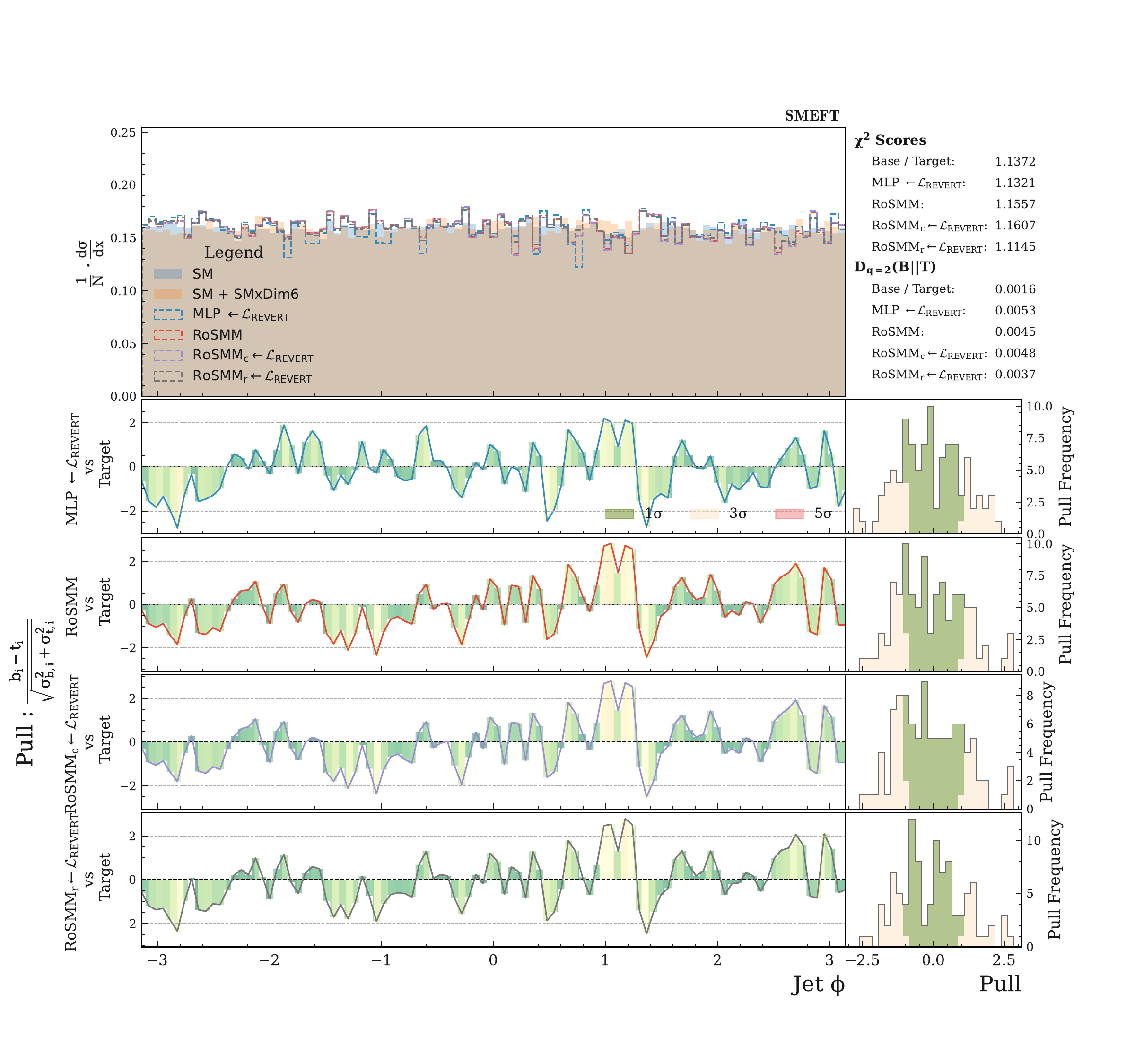}
\includegraphics[scale=0.21]{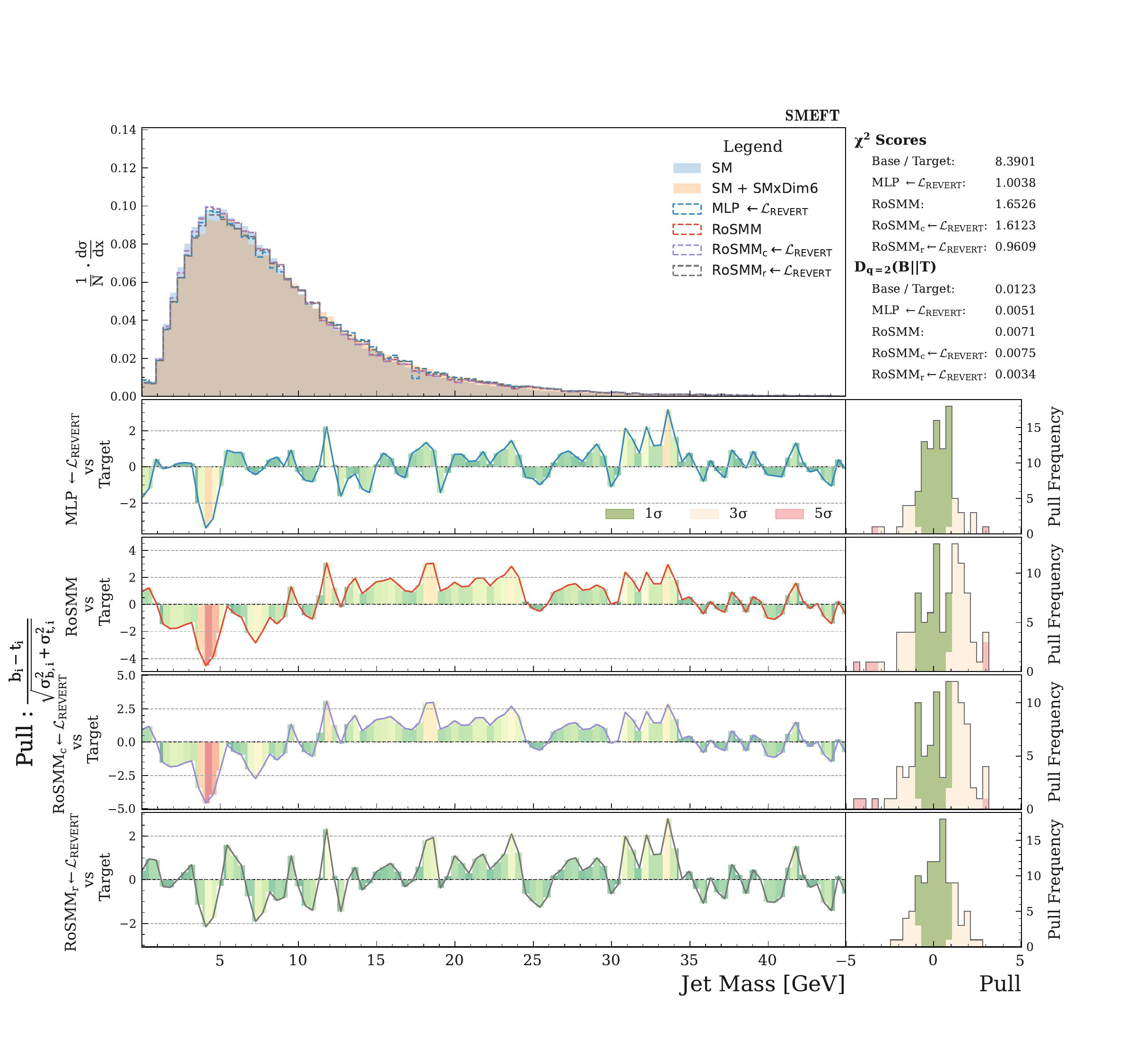}
\caption{Reweighting closure plots for the Jet features $(p_T,\eta,\phi,m)$ where the reference (Standard Model) distribution is mapped to the target (SMEFT) distribution using the different density ratio estimation models.}
\end{figure}

\begin{figure}[H]
\centering
\includegraphics[scale=0.21]{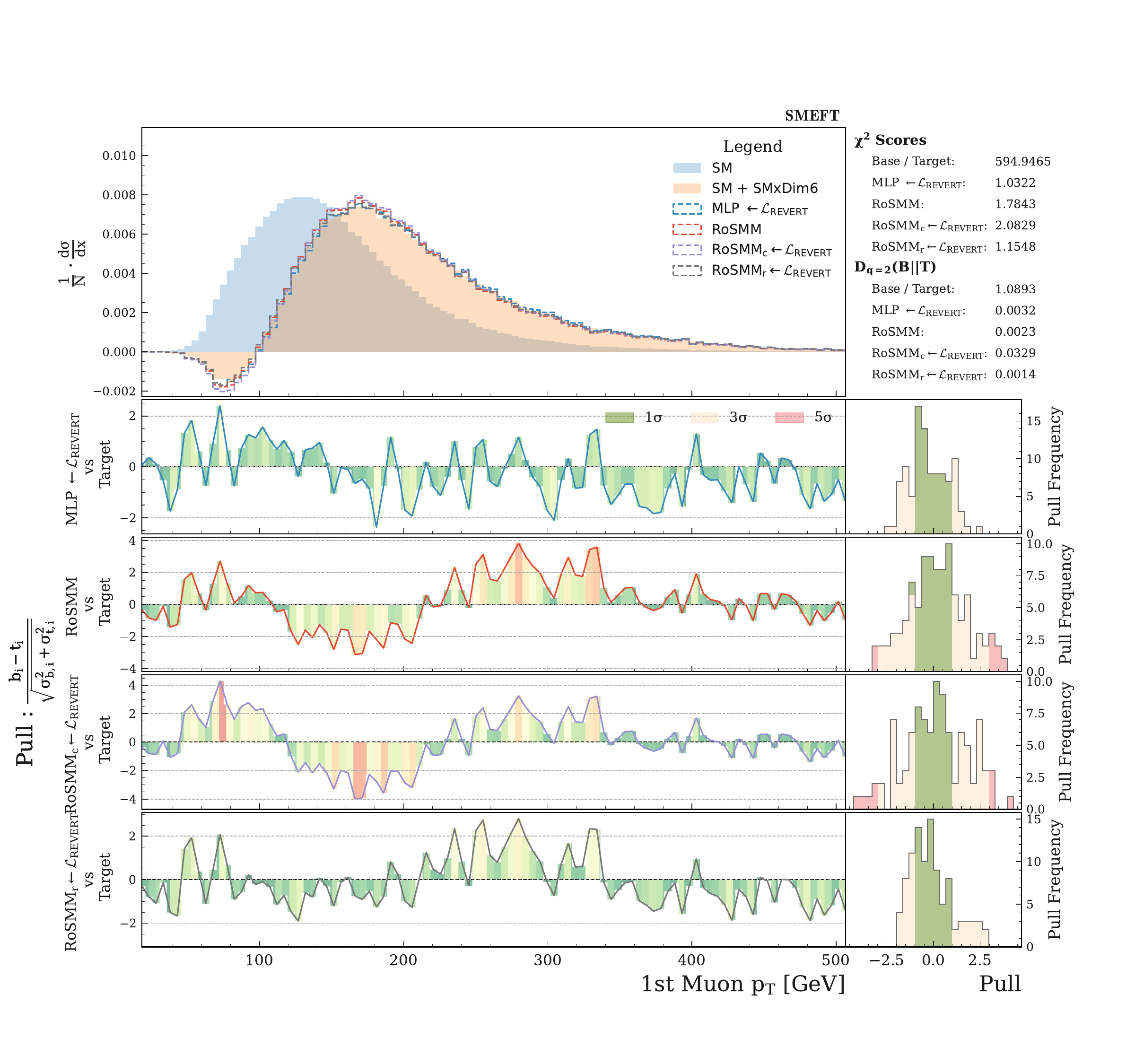}
\includegraphics[scale=0.21]{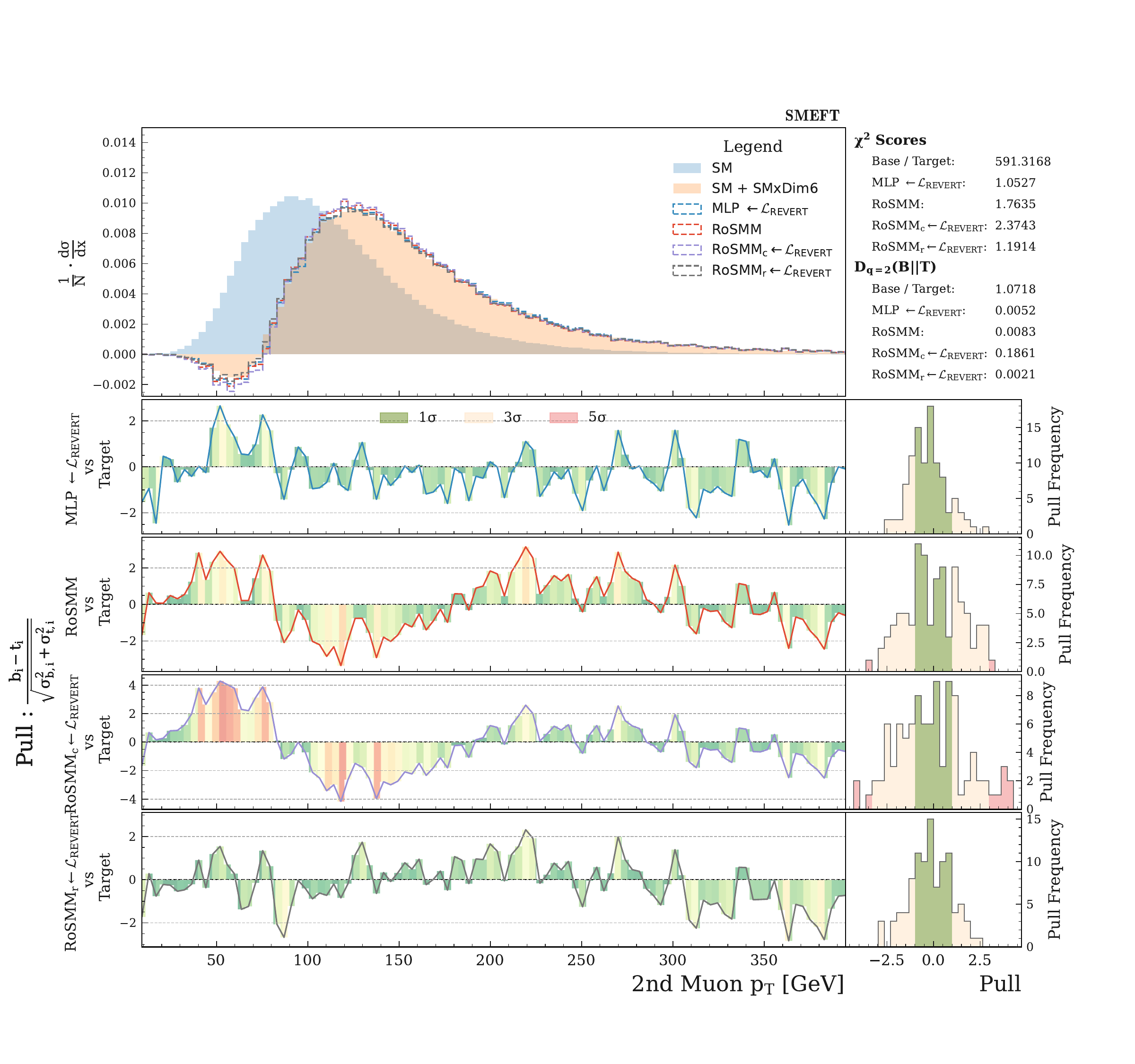}
\includegraphics[scale=0.21]{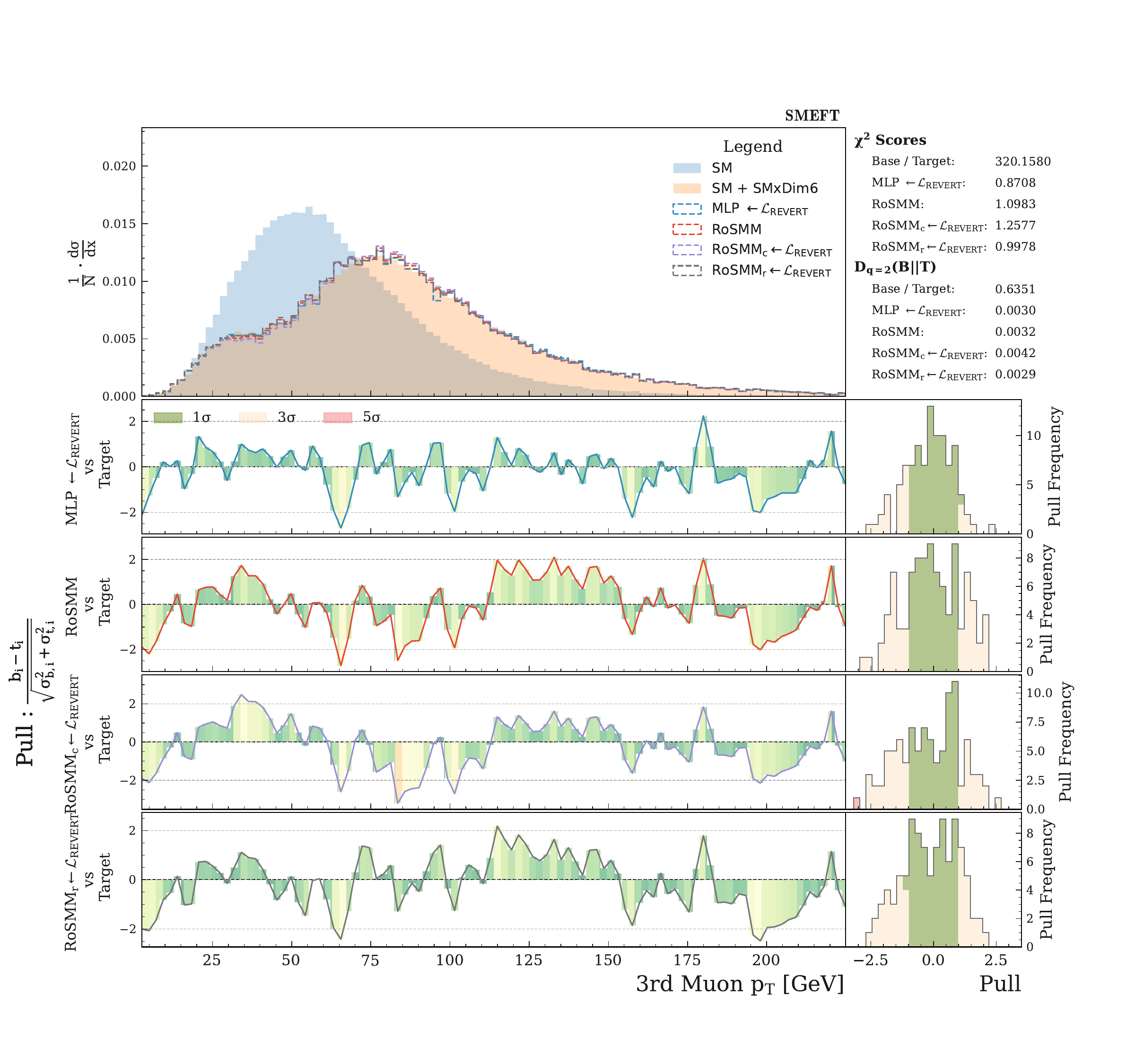}
\includegraphics[scale=0.21]{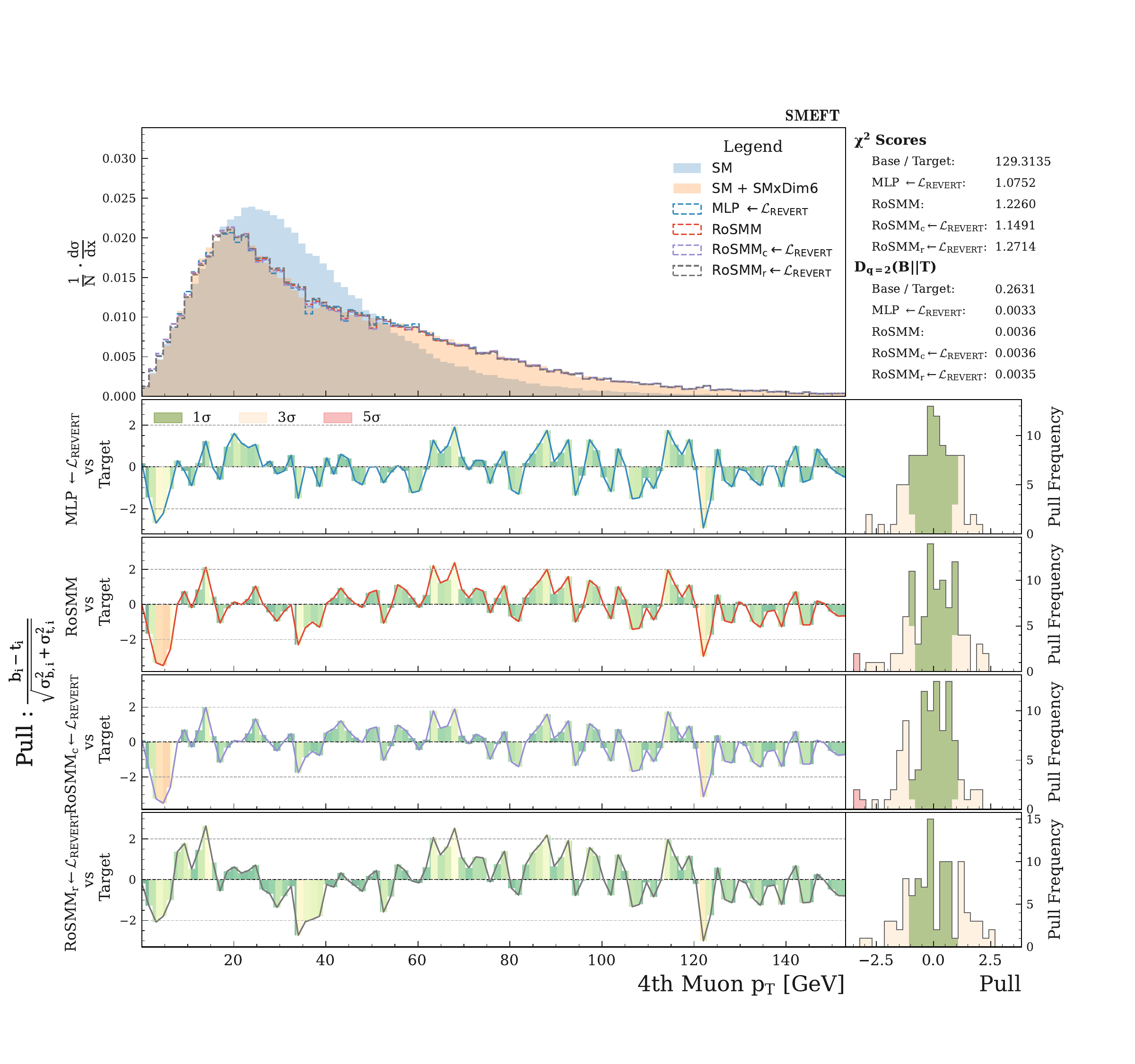}
\caption{Reweighting closure plots for the four muon $p_T$ features where the reference (Standard Model) distribution is mapped to the target (SMEFT) distribution using the different density ratio estimation models.}
\end{figure}

\begin{figure}[H]
\centering
\includegraphics[scale=0.21]{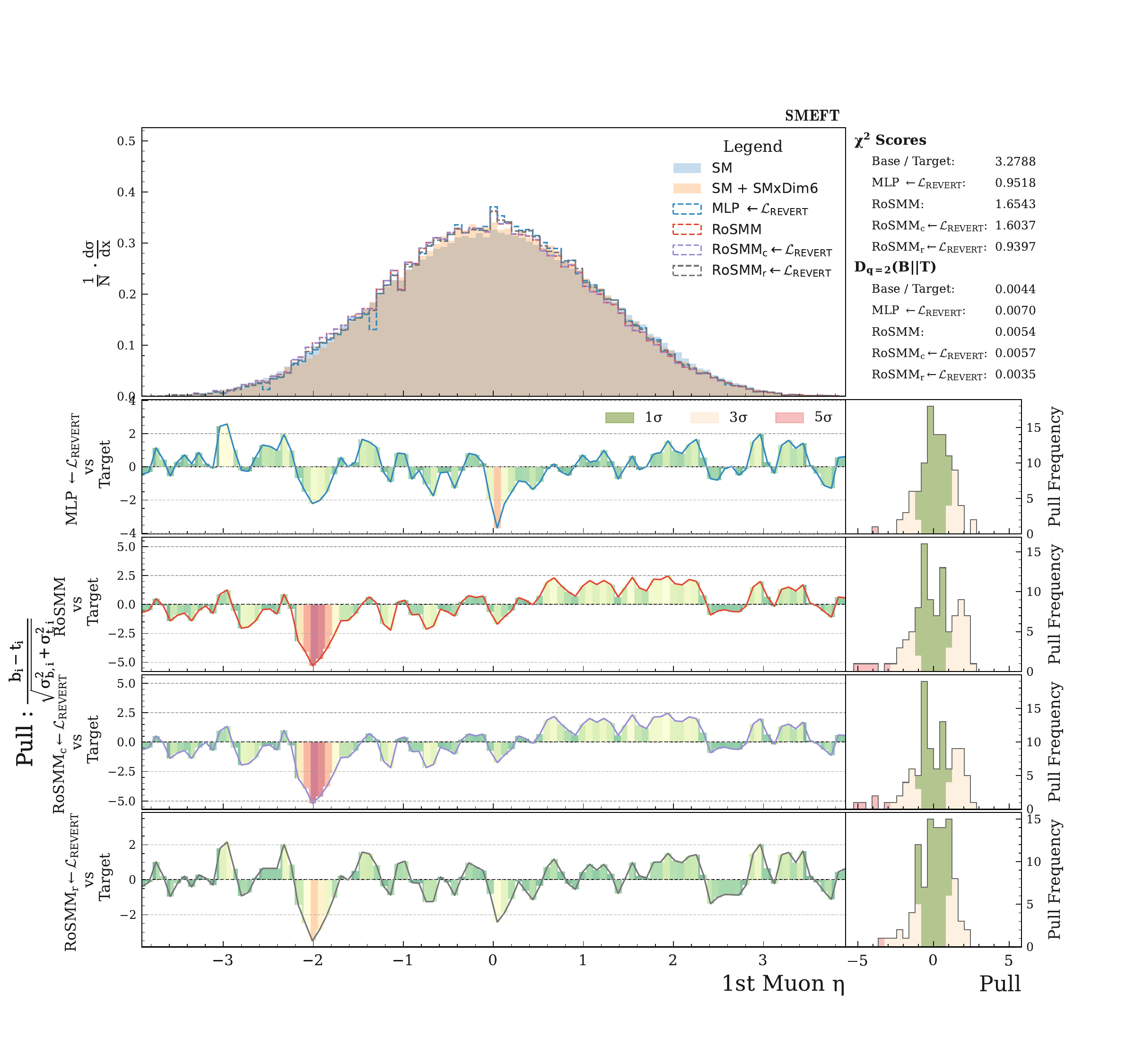}
\includegraphics[scale=0.21]{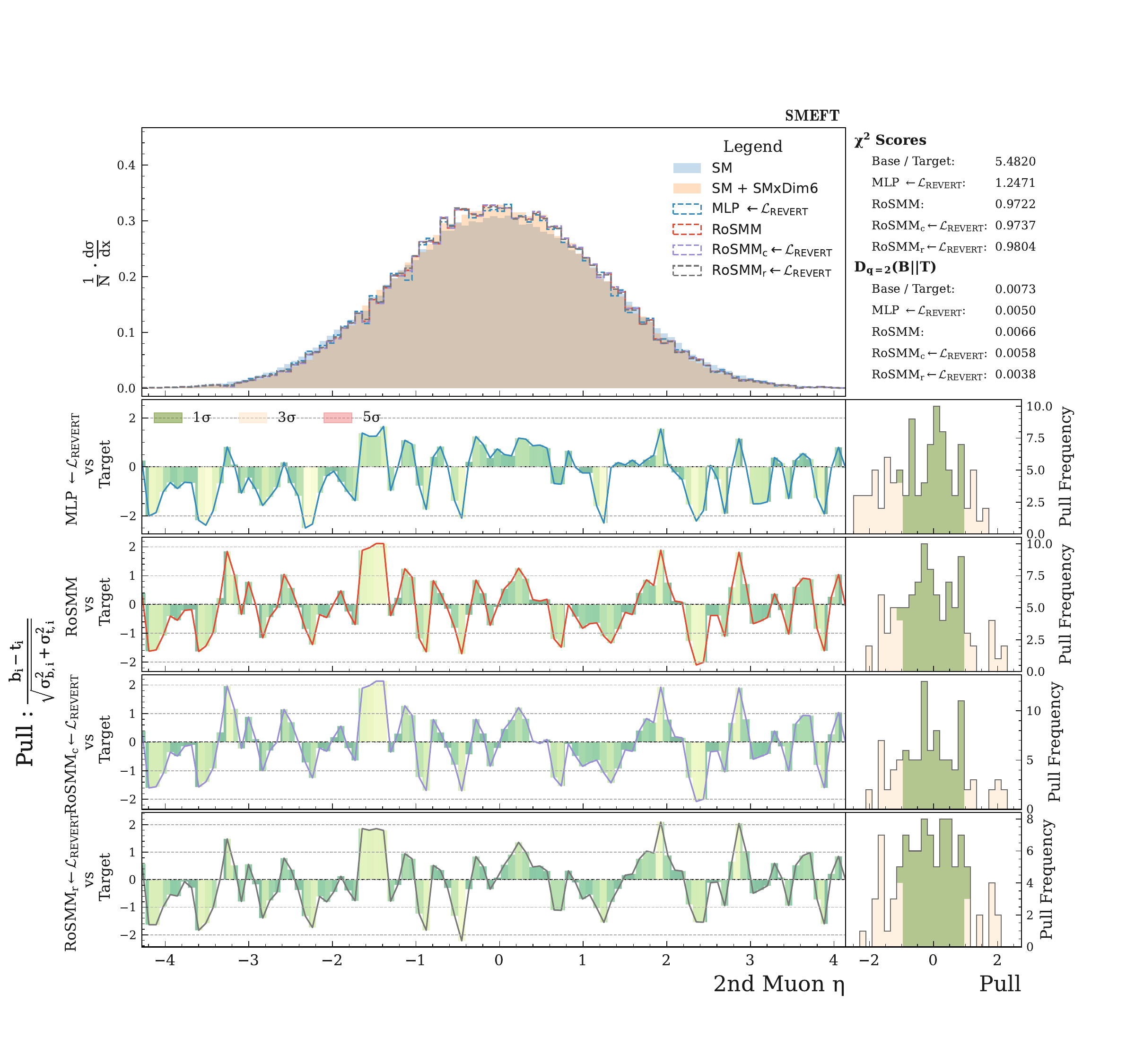}
\includegraphics[scale=0.21]{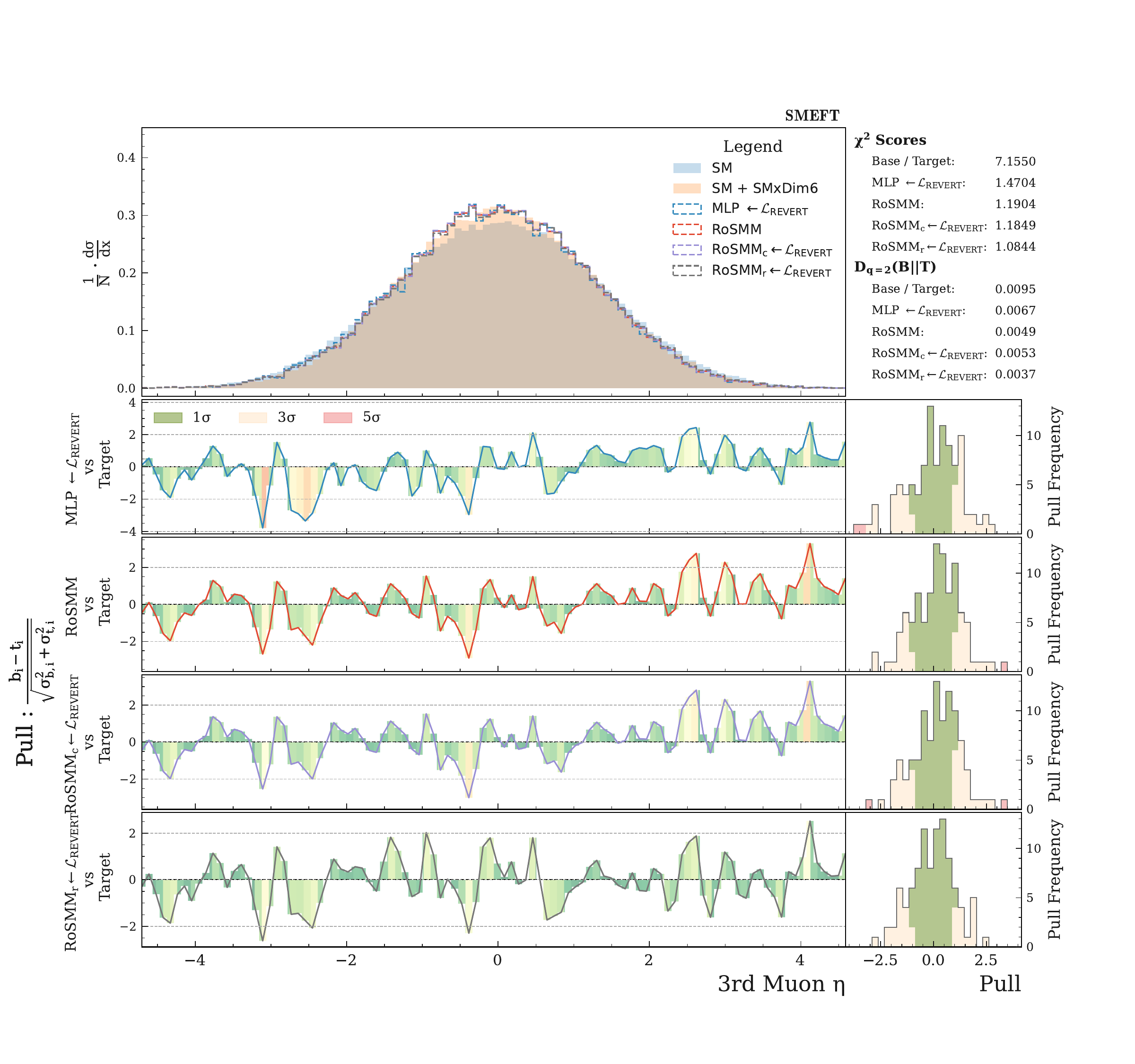}
\includegraphics[scale=0.21]{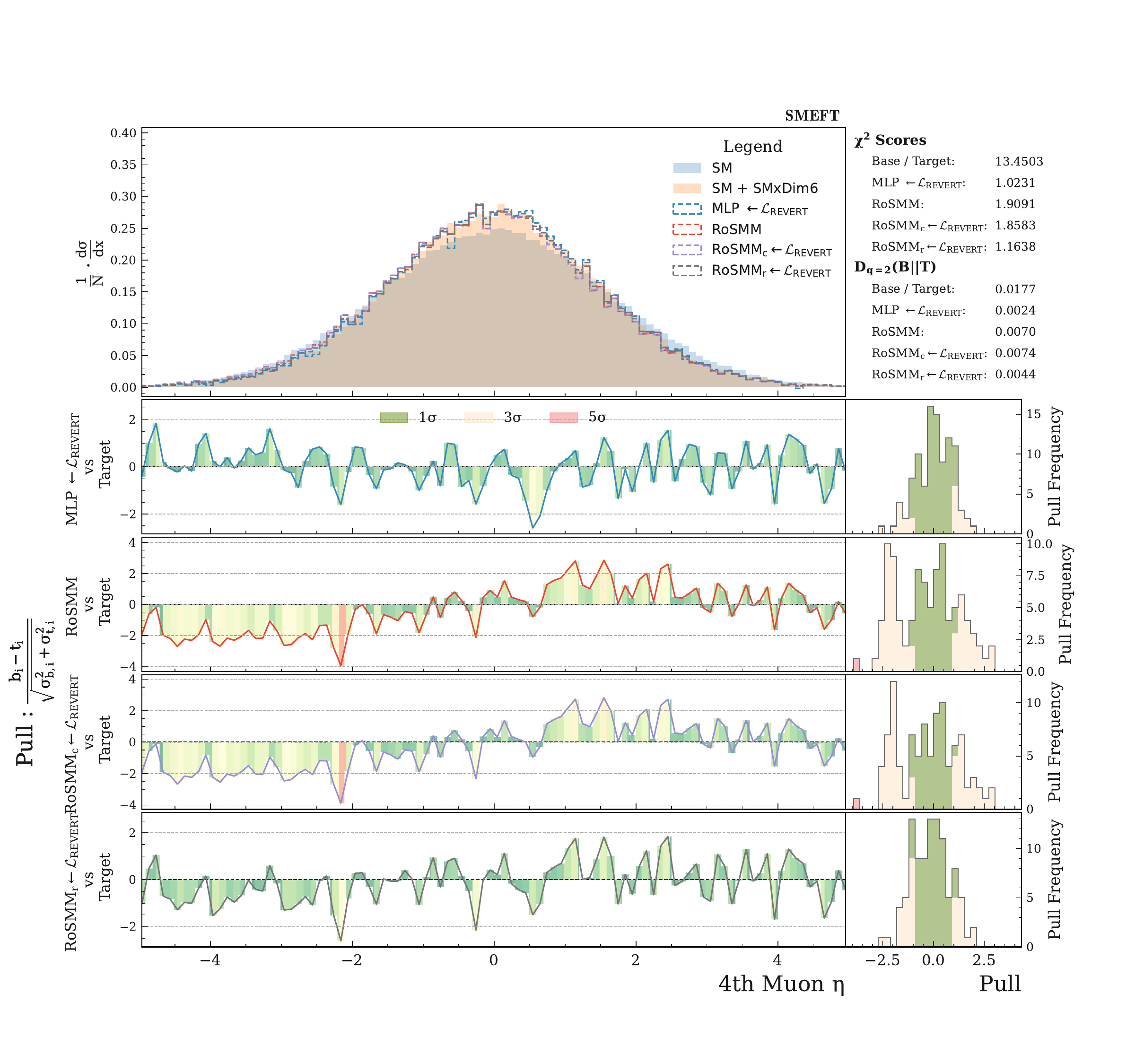}
\caption{Reweighting closure plots for the four muon $\eta$ features where the reference (Standard Model) distribution is mapped to the target (SMEFT) distribution using the different density ratio estimation models.}
\end{figure}

\begin{figure}[H]
\centering
\includegraphics[scale=0.21]{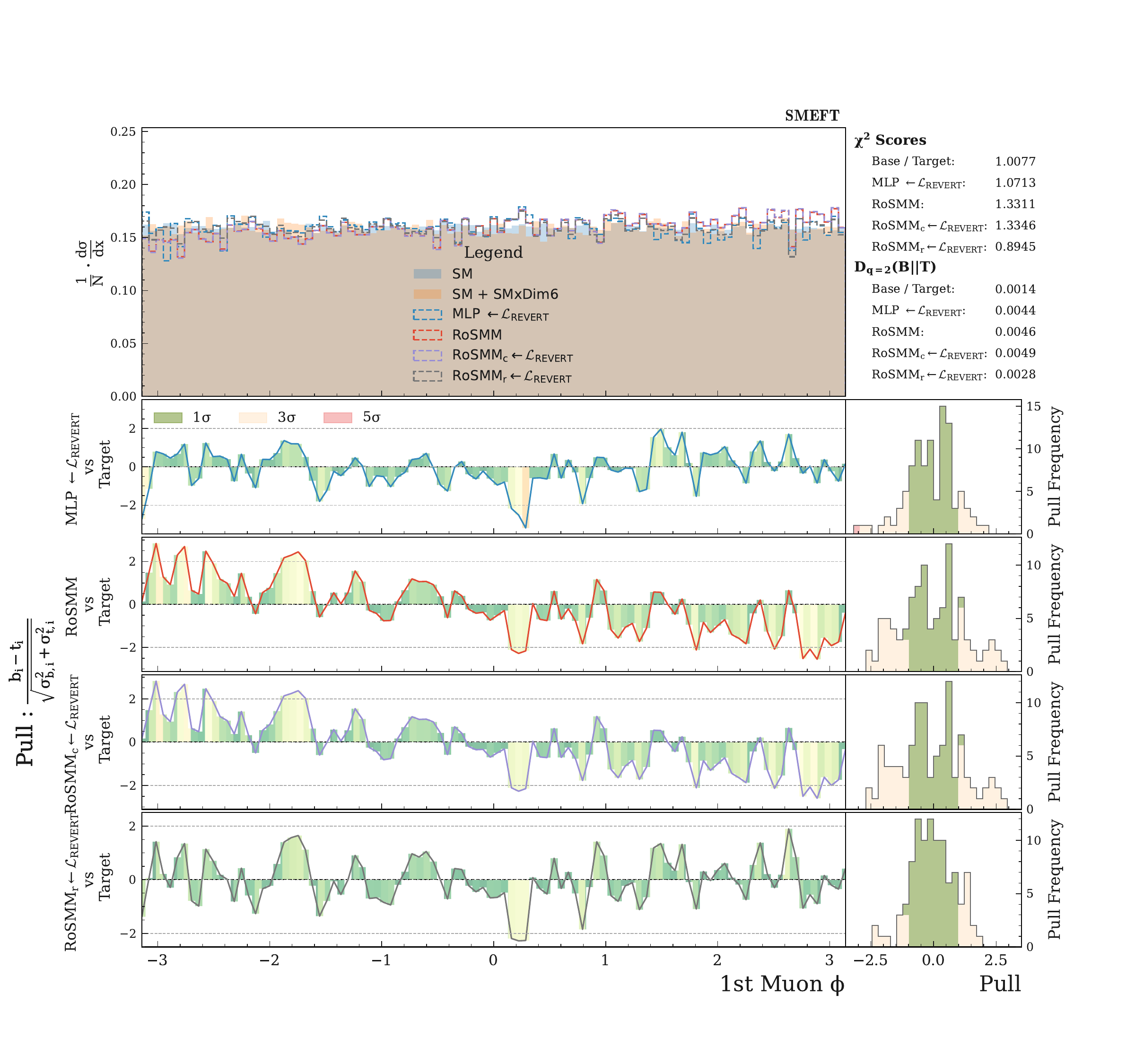}
\includegraphics[scale=0.21]{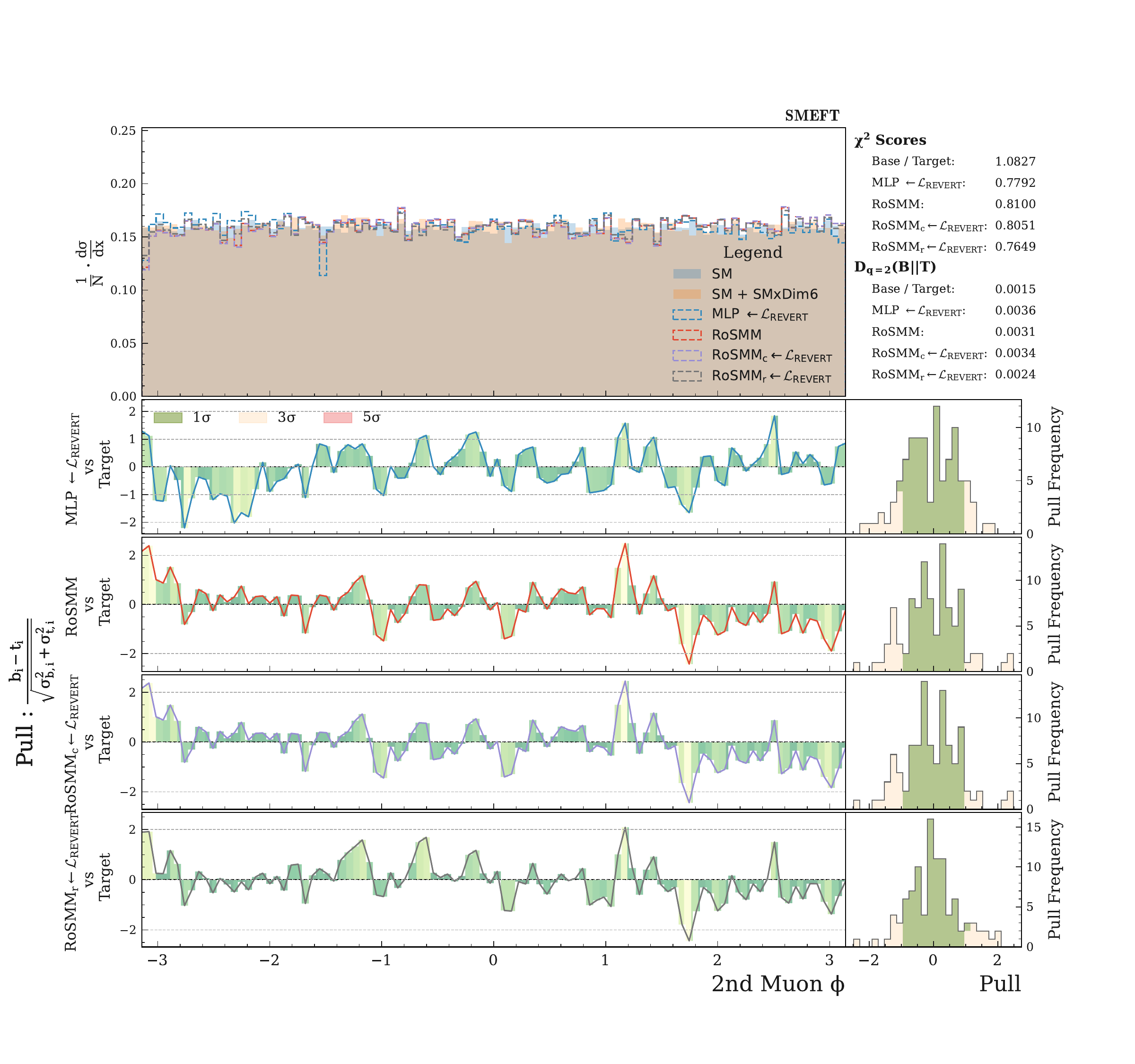}
\includegraphics[scale=0.21]{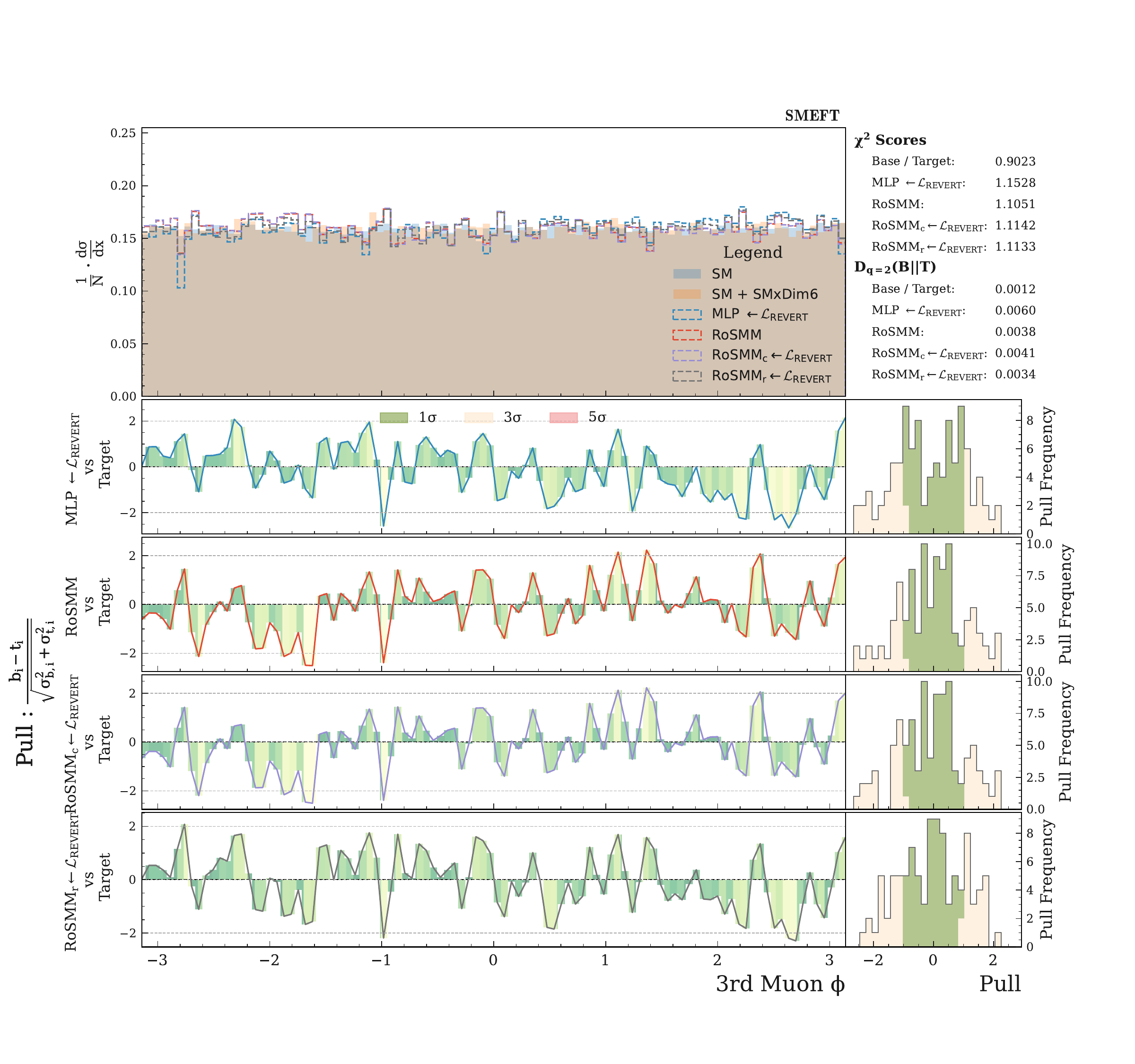}
\includegraphics[scale=0.21]{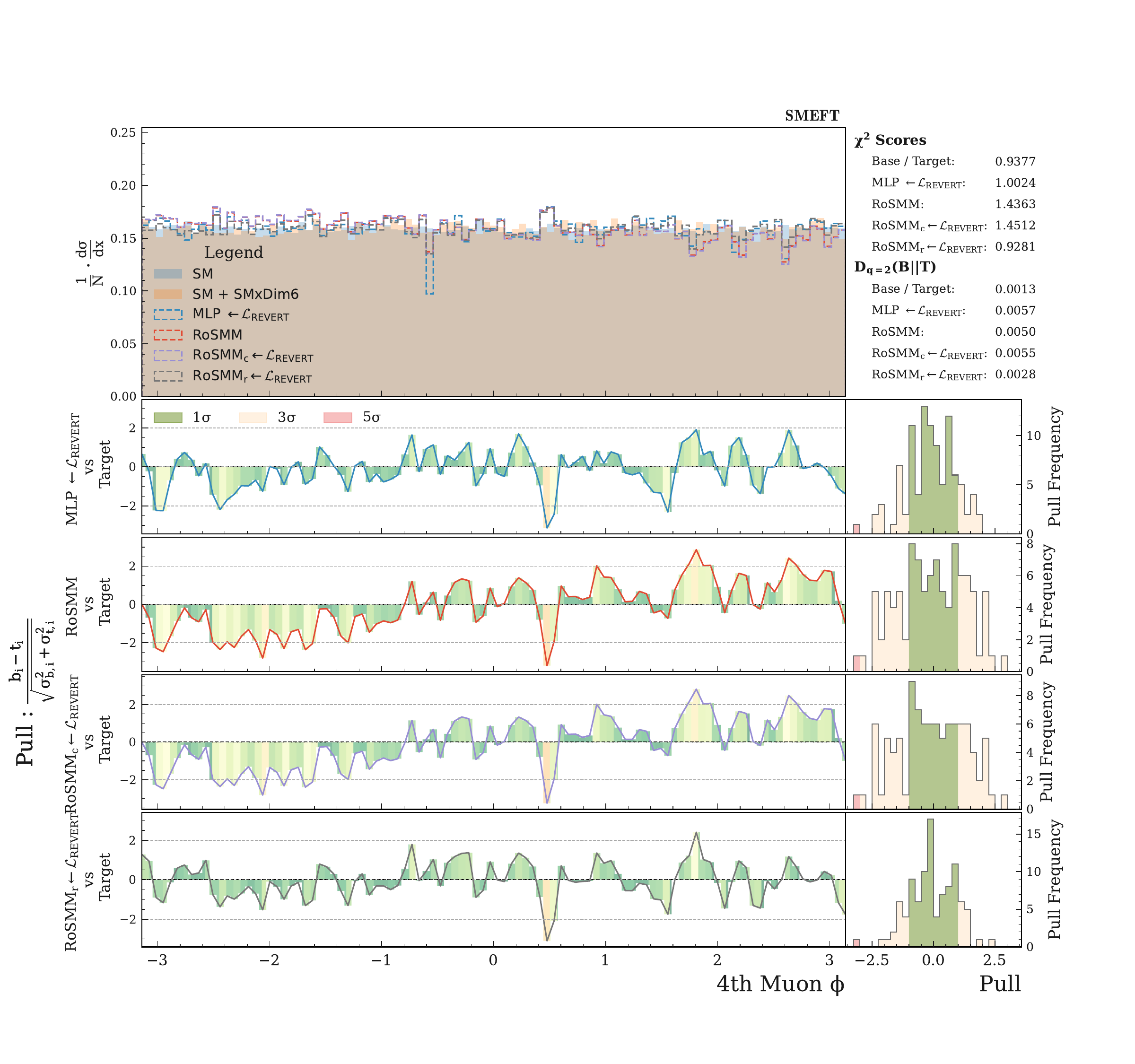}
\caption{Reweighting closure plots for the four muon $\phi$ features where the reference (Standard Model) distribution is mapped to the target (SMEFT) distribution using the different density ratio estimation models.}
\end{figure}

\begin{figure}[H]
\centering
\includegraphics[scale=0.21]{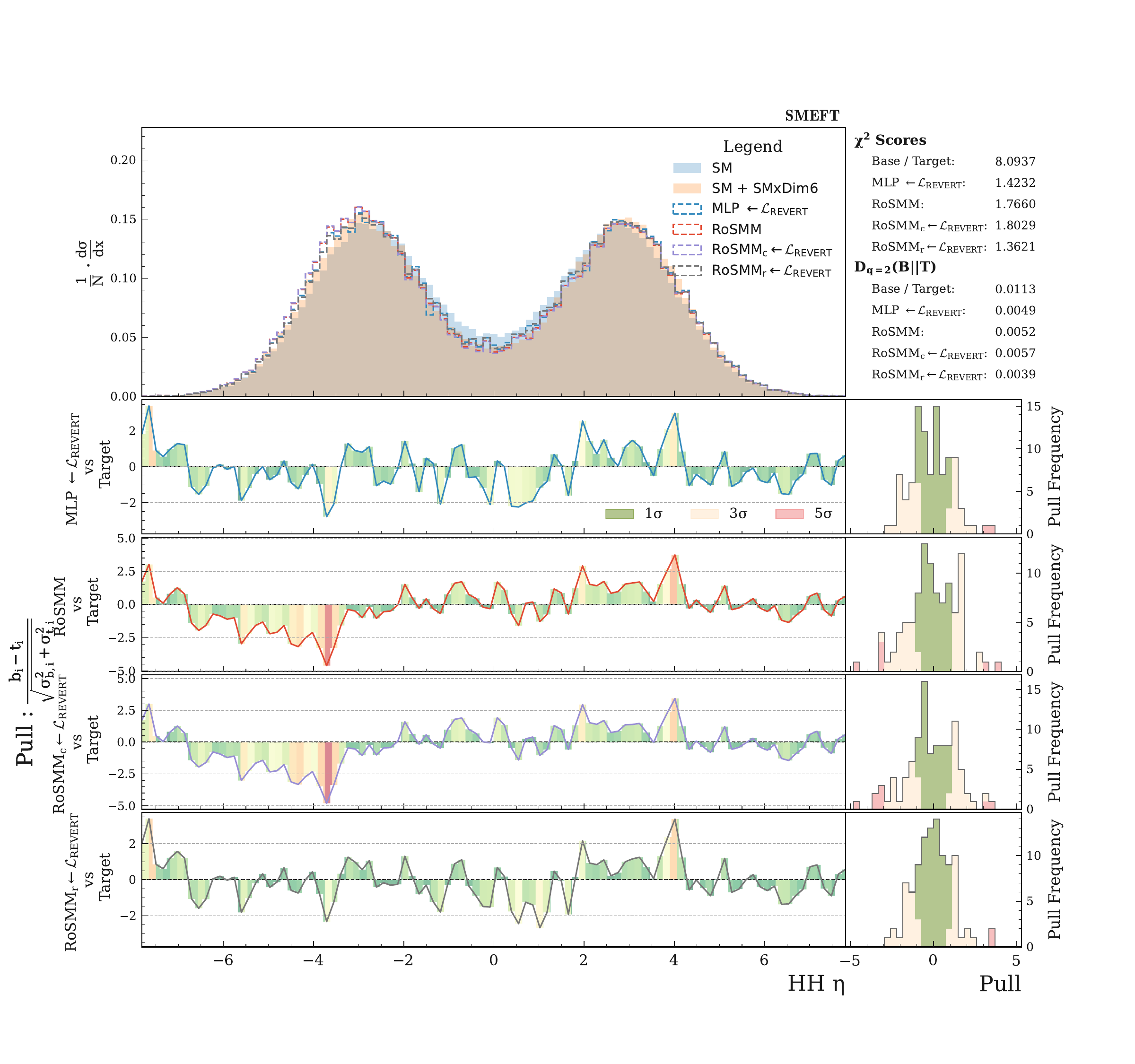}
\includegraphics[scale=0.21]{images/SMEFT/SMEFT_hh_eta_closure.pdf}
\includegraphics[scale=0.21]{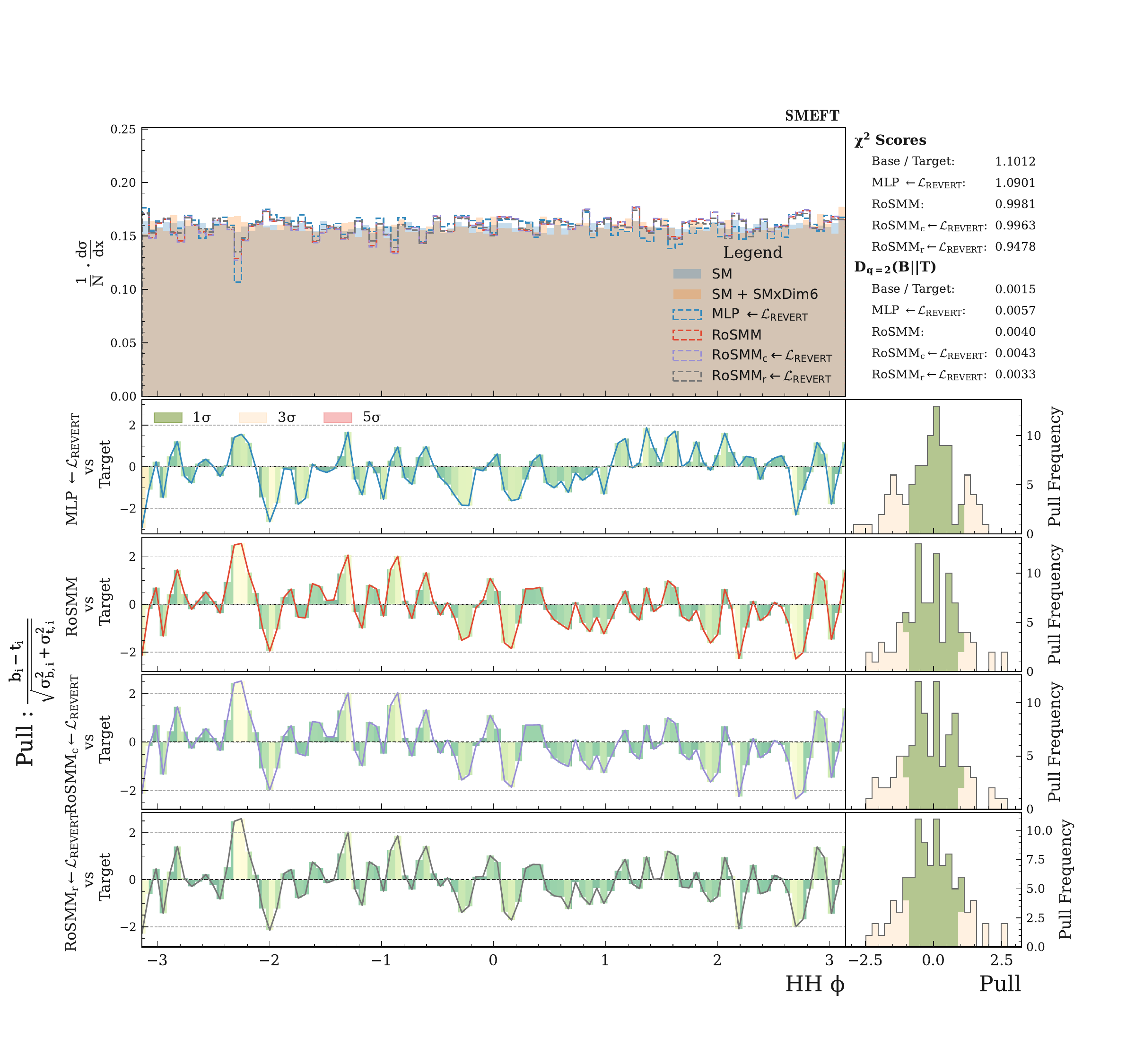}
\includegraphics[scale=0.21]{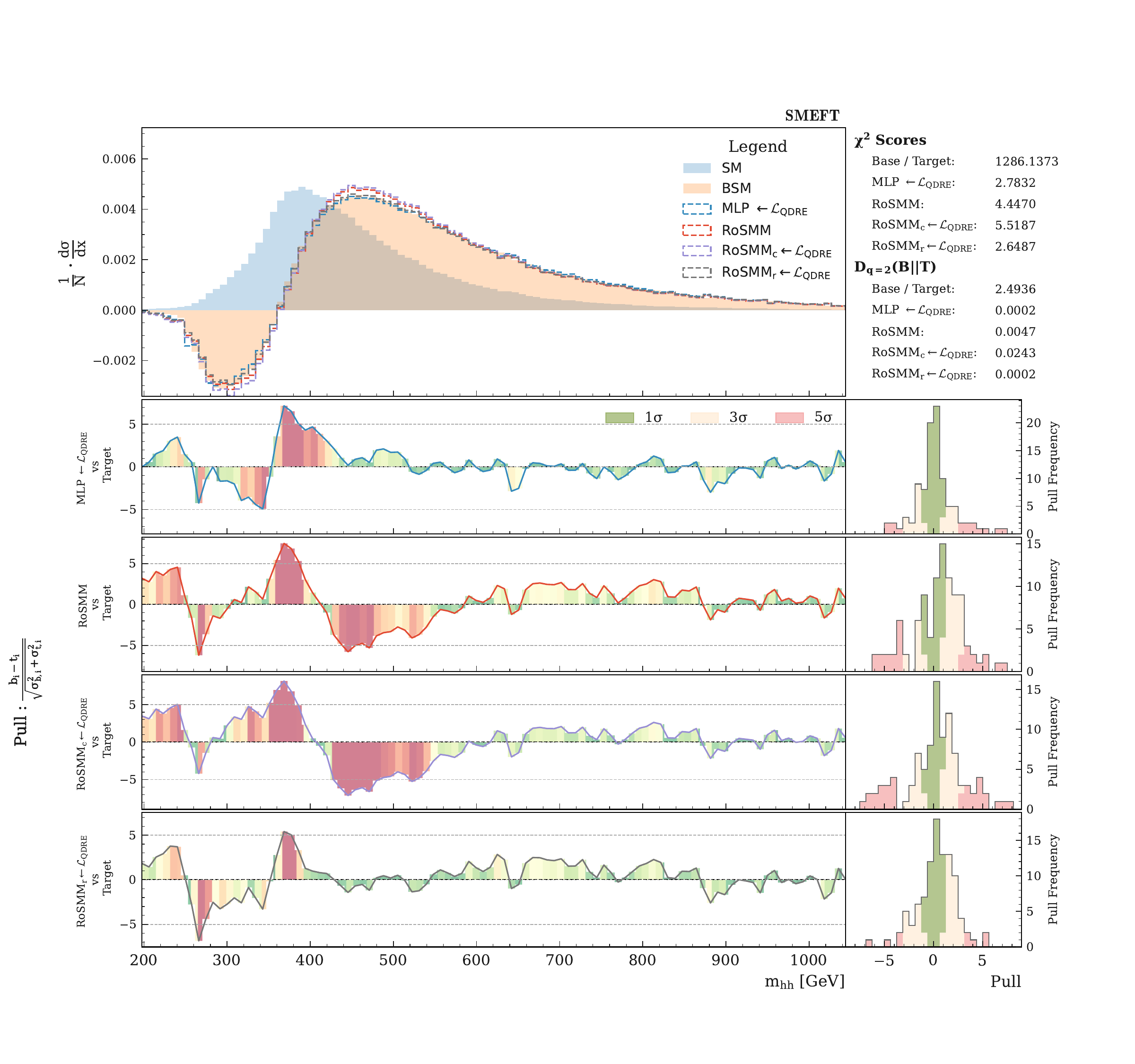}
\caption{Reweighting closure plots for the di-Higgs features $(p_T,\eta,\phi)$ where the reference (Standard Model) distribution is mapped to the target (SMEFT) distribution using the different density ratio estimation models.}
\end{figure}

\clearpage

\end{document}